\documentclass{article}

\usepackage{microtype}
\usepackage{graphicx}
\usepackage{subfigure}
\usepackage{booktabs} 

\usepackage{hyperref}

\usepackage[accepted]{icml2024}

\usepackage[utf8]{inputenc} 
\usepackage[T1]{fontenc}    
\usepackage{hyperref}       
\usepackage{url}            
\usepackage{booktabs}       
\usepackage{amsfonts}       
\usepackage{nicefrac}       
\usepackage{microtype}      
\usepackage{xcolor}         

\usepackage{algorithm}
\usepackage{algorithmic}
\renewcommand{\algorithmicrequire}{\textbf{Input:}} 
\renewcommand{\algorithmicensure}{\textbf{Output:}}  

\usepackage{amsmath}
\usepackage{graphics}
\usepackage{epsfig}
\usepackage{multirow}
\usepackage{xspace}
\usepackage{subfigure}
\usepackage{fancyhdr}
\usepackage{microtype}
\usepackage{color}
\usepackage{colortbl}
\usepackage{xcolor}
\definecolor{mygray}{gray}{.85}
\definecolor{myyellow}{RGB}{204,102,0}
\definecolor{myred}{RGB}{204,0,102}
\definecolor{mypurple}{RGB}{102,0,204}
\definecolor{maroon}{cmyk}{0,0.87,0.68,0.32}
\definecolor{myblue}{RGB}{227,227,240}

\usepackage{graphics}
\usepackage{amssymb}
\usepackage{amsthm}
\usepackage{mathrsfs}
\usepackage{booktabs}
\usepackage{algorithm}
\usepackage{algorithmic}
\usepackage{arydshln}
\usepackage{siunitx}
\usepackage{appendix}
\usepackage{enumitem}
\usepackage{wrapfig}

\theoremstyle{plain}

\theoremstyle{definition}

\theoremstyle{remark}

\newcommand{\model}{EnzyGen\xspace}
\newcommand{\dataset}{EnzyBench\xspace}
\newcommand{\layer}{NAEL\xspace}

\icmltitlerunning{Generative Enzyme Design Guided by Functionally Important Sites and Small-Molecule Substrates}

\begin{document}


\twocolumn[
\icmltitle{Generative Enzyme Design Guided by Functionally Important Sites and Small-Molecule Substrates}



\icmlsetsymbol{equal}{*}

\begin{icmlauthorlist}
\icmlauthor{Zhenqiao Song}{yyy}
\icmlauthor{Yunlong Zhao}{}
\icmlauthor{Wenxian Shi}{}
\icmlauthor{Wengong Jin}{}
\icmlauthor{Yang Yang}{eee}
\icmlauthor{Lei Li}{yyy}
\end{icmlauthorlist}

\icmlaffiliation{yyy}{Language Technologies Institute, Carnegie Mellon University.}
\icmlaffiliation{eee}{Department of Chemistry and Biochemistry, University of California Santa Barbara}

\icmlcorrespondingauthor{Lei Li}{leili@cs.cmu.edu}
\icmlcorrespondingauthor{ Yang Yang}{yang@chem.ucsb.edu}

\icmlkeywords{Machine Learning, ICML}

\vskip 0.3in
]



\printAffiliationsAndNotice{We have added \model-1.5 to this version. This new model undergoes initial pretraining with masked language modeling before proceeding through the standard \model training pipeline.}  

\begin{abstract}
Enzymes are genetically encoded biocatalysts capable of accelerating chemical reactions. How can we automatically design functional enzymes?
In this paper, we propose \model, an approach to learn a unified model to design enzymes across massive functional families. Our key idea is to generate an enzyme's amino acid sequence and their three-dimensional~(3D) coordinates based on functionally important sites and substrates corresponding to a desired catalytic function. These  sites are automatically mined from enzyme databases.
\model consists of a novel interleaving network of attention and neighborhood equivariant layers, which captures both long-range correlation in an entire protein sequence and local influence from nearest  amino acids in 3D space. To learn the generative model, we devise a joint training objective, including a sequence generation loss, a position prediction loss and an enzyme-substrate interaction loss. We further construct \dataset, a dataset with 3157 enzyme families, covering all available enzymes within the protein data bank~(PDB).
Experimental results show that our \model consistently achieves the best performance across all 323 testing families, surpassing the best baseline by 10.79\% in terms of substrate binding affinity. 
These findings demonstrate \model's superior capability in designing well-folded and effective enzymes binding to specific substrates with high affinities. The code, model and dataset are released at \url{https://github.com/LeiLiLab/EnzyGen}.

\end{abstract}

\section{Introduction}
\label{introduction}

Enzymes are biological catalysts to accelerate challenging chemical reactions underlying a range of biological processes. They enjoy widespread applications in the production of pharmaceuticals~\cite{wu2012bioengineering}, specialty chemicals~\cite{carbonell2018selenzyme} and biofuels~\cite{liao2016fuelling}.
In enzymatic reactions, a substrate is a  molecule being converted by the enzyme catalyst. By binding with and acting on specific substrates, enzymes allow for dramatically accelerated rates in the transformation of their substrates~\cite{bar2011moderately}. Designing enzymes that can bind to specific substrates is a critical yet challenging problem.



Recently, deep learning methods are promising in protein design~\cite{huang2016coming,pearce2021deep}. Existing deep learning approaches for functional protein design fall into three categories: (a) generative models for protein sequence design guided by fitness landscape~\cite{brookes2019conditioning,rives2021biological,ren2022proximal,song2023importance,wang2023self}; (b) structure to sequence generation based on a targeted protein backbone~\cite{ingraham2019generative,dauparas2022robust,zheng2023structure}; (c) co-design of a backbone structure and a protein sequence that encodes it~\cite{anishchenko2021novo,wang2022scaffolding,shi2022protein,yeh2023novo}. However, these methods face many limitations on enzyme design. First, fitness-guided approaches are limited by the lack of fitness data for the vast majority of enzyme families. Second, the structures of many enzymes remain unknown~\cite{binz2005engineering,fischman2018computational}. Third, prior approaches do not model substrates in enzyme design process. Finally, there is no unified model applicable to massive and diverse enzyme families.

In this paper, we aim to develop a unified generative model to design functional enzymes across thousands of enzyme families.
The key design idea lies in the notion that the biological function of an enzyme is enabled by a subset of residues~(amino acids), known as functionally important sites~\cite{bickel2002finding,chakrabarti2007analysis,wang2022scaffolding}, and that an ideal enzyme should be able to bind its substrate(s) in enzymatic reactions. 
Therefore, we formulate the enzyme design problem as jointly generating the enzyme sequence and backbone structure  given an enzyme's functional category, automatically mined functionally important sites, and associated substrates in the catalytic reactions of the enzyme.

To this end, we propose \model, a unified model to co-design enzyme sequence and backbone structure. \model comprises an enzyme modeling module and a substrate representation module. We devise neighborhood attentive equivariant layers~({\layer}s) to jointly model enzyme amino acid sequence and backbone structure coordinates.
Each \layer consists of a global attention sub-layer and a neighborhood equivariant sub-layer. The global attention sub-layer captures correlations among all residues within an entire enzyme sequence, while the neighborhood equivariant sub-layer updates residue representations and coordinates based on nearest neighbors in the 3D space. This architectural design facilitates information exchange with varying levels of granularity, promoting a comprehensive interaction among different residues. The substrate representation module also uses neighborhood equivariant layers to model a given substrate. Furthermore, \model learns embeddings for enzyme tags corresponding to the BRENDA database
to enable transfer among enzymes within the same family branch. To train \model, we design a joint training objective consisting of an amino acid type prediction loss, a residue coordinate reconstruction loss and an enzyme-substrate interaction loss.

Our contributions are listed as follows: 
\begin{itemize}[nosep,leftmargin=2.6em]
    \item We propose \model to jointly generate enzyme sequence and backbone structure. \model is the first unified enzyme design model across thousands of enzyme families.
    \item We create \dataset, a comprehensive benchmark for training and evaluating enzyme design. It includes $101,974$ PDB entries across $3,157$ enzyme families, and a set of three metrics to evaluate enzyme quality prior to wet-lab experiments. \dataset includes a test set with $323$ families for various function validation. 
    \item We conduct experiments to evaluate \model and existing functional protein design methods. \model outperforms the best prior baseline by 10.79\% in terms of substrate binding affinity. We show that enzymes designed by \model exhibit the best average enzyme-substrate ESP score~\cite{kroll2023general} of 0.65, substrate binding affinity of -9.44, and AlphaFold2 pLDDT~\cite{jumper2021highly} of 87.45 across 323 families (BRENDA enzyme classification fourth-level categories). 
\end{itemize}

\section{Related Work}
\label{related_work}
\textbf{Methods for Functional Protein Design.}
Functional protein design has been studied with a wide variety of methods. 
Some works focus on designing functional proteins guided by fitness landscape, including searching~\cite{brookes2018design,brookes2019conditioning,kumar2020model,das2021accelerated,hoffman2022optimizing,melnyk2021benchmarking,anishchenko2021novo,ren2022proximal} or directly generating~\cite{jain2022biological,song2023importance} protein sequences through deep generative models. Another category of methods concentrates on designing protein sequences capable of folding into backbone structures to fulfill specific functions~\cite{wang2018computational,strokach2020fast,dauparas2022robust,hsu2022learning,sumida2023improving}. Owing to the availability of large-scale data, some models are trained on extensive protein sequences from diverse families~\cite{ferruz2022protgpt2} or incorporate additional control tags for enhanced modeling~\cite{madani2023large}.
On the path from sequence to function, the role of structure emerges as a crucial intermediary. Consequently, some studies focus on the design of functional structures~\cite{trippe2022diffusion,watson2023novo}. Recognizing the inherently complex interplay between protein sequence and structure, another set of works relies on the simultaneous generation of a new backbone structure and an amino acid sequence that folds into it~\cite{anishchenko2021novo,wang2022scaffolding,shi2022protein,yeh2023novo,song2023functional}.
Compared to the design of other functional proteins, functional enzyme design is largely underdeveloped. Some of the above methods have been applied to specific enzyme families, such as generating satisfactory enzyme sequences using deep generative models~\cite{detlefsen2022learning,lin2022novo,giessel2022therapeutic}. However, no previous model is capable of generating enzymes in all families (over 3,000).

\begin{figure*}
  \centering
  \includegraphics[width=17.0cm]{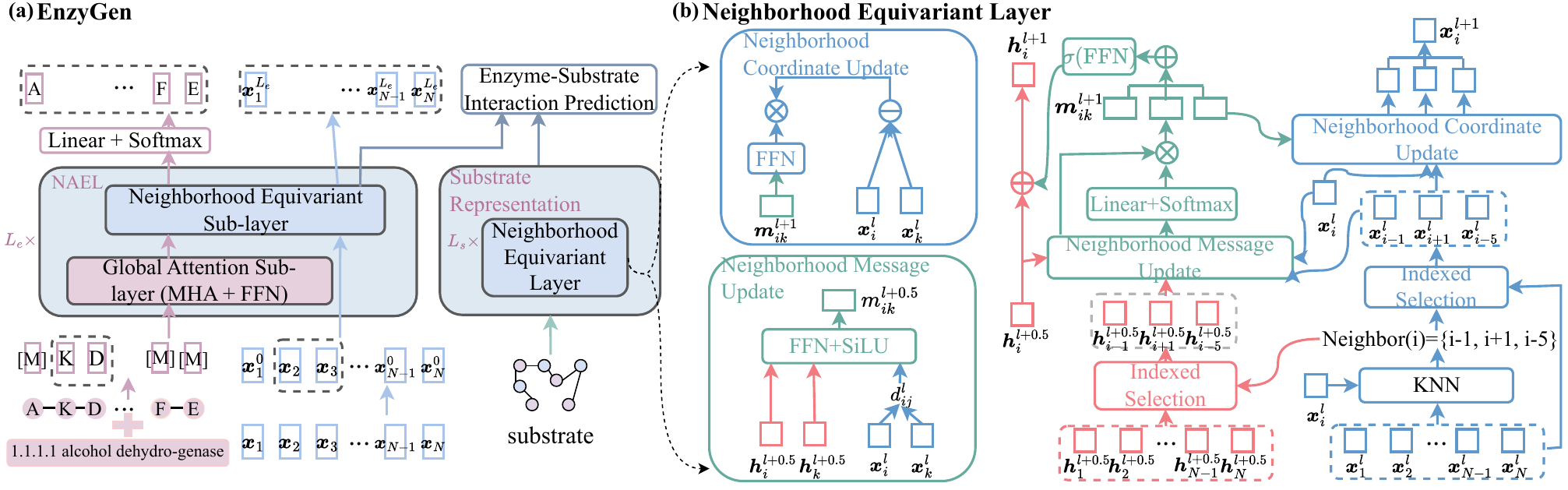}
  \vspace{-0.6em}
  \caption{
  (a) \model architecture, consisting of an enzyme modeling module~(left) and a substrate representation module~(right). The enzyme modeling module aims to generate the enzyme sequence and backbone structure, and the substrate representation module targets at predicting if an enzyme can bind to a substrate. The dashed box in enzyme input denotes functionally important sites, while other sites need to be generated. [M] denotes mask token. ``1.1.1.1" denotes the fourth-level enzyme class in the BRENDA enzyme classification (EC) tree. 
  (b) Neighborhood equivariant layer: neighborhood message update~(in green), neighborhood coordinate update (in blue) and neighborhood node feature update (in red). Indexed selection is choosing $\boldsymbol{x}_j$ (or $\boldsymbol{h}_j$) where $j^{th}$ residue is in the $K$-nearest neighbors of $i^{th}$ residue.}
  \label{Fig: model}
\end{figure*}

\textbf{Functionally Important Site Discovery in Enzymes.} As our method conditions enzyme design on automatically mined functionally important sites, crucial in determining their structures or functions~\cite{tristem2000molecular}, it is necessary to revisit research on identifying functionally important sites in enzymes. Several databases, such as PROSITE~\cite{hulo2006prosite}, Gene Ontology~\cite{ashburner2000gene} and InterPro~\cite{hunter2009interpro} identify and annotate protein functional sites based on sequentially conserved residues and information extracted from experimental studies and literature searches. Some efforts have also been undertaken to derive functional insights using 3D structural information~\cite{fetrow1998method,fetrow2001genomic,ivanisenko2005pdbsite}. Other investigators have predicted functional sites using multiple sequence alignments~\cite{armon2001consurf,panchenko2004prediction,bray2009sitesidentify,hosseini2022pithia}.

\section{Proposed Method: \model}
\label{methods}

An enzyme consists of a chain of amino acids~(also called residues) connected by peptide bonds, which folds into a proper 3D structure.
Let $\mathcal{A}$ be the set of 20 common amino acids.
We denote the sequence of a $N$-residue enzyme by $\boldsymbol{s}=\{s_1, s_2, ..., s_N\} \in \mathcal{A}^N$ and their $C_{\alpha}$ coordinates by $\boldsymbol{x}=[\boldsymbol{x}_1, \boldsymbol{x}_2, ..., \boldsymbol{x}_N]^T \in \mathbb{R}^{N\times3}$.
For residue type $s_i\in \mathcal{A}$, where $i \in \{1, 2, ..., N\}$, we denote its one-hot encoding as $\boldsymbol{s}_i=\text{onehot}(s_i)$ and the functionally important site index set as $\mathcal{M}$. 
We denote a substrate as $\mathcal{V}=\{v_1,...,v_m\}$, where $m$ is the number of its atoms, and $v_j=(\boldsymbol{h}_j, \boldsymbol{x}_j)$ denotes an atom where $j \in \{1, 2, ..., m\}$. Here, $\boldsymbol{h}_j$ represents pre-computed chemical features, and $\boldsymbol{x}_j\in \mathbb{R}^3$ is the corresponding coordinate. Each substrate is paired with a binary label $y$ indicating whether the enzyme can bind with it.
Inspired by multilingual machine translation models in natural language processing, which employ language tags to guide model generation in specific target languages~\cite{johnson2017google,liu2020multilingual}, we similarly use enzyme family tags to facilitate enzyme design with desired functions. To provide a systematical and professional enzyme family classification, we leverage the enzyme classification~(EC) tree in the BRENDA database~\cite{schomburg2004brenda}. For example, the EC tag 1.1.1.1 denotes enzyme family of alcohol dehydrogenase.

The problem studied in this paper can be formulated as follows: given four-level EC tag $c=\{c_1, c_2, c_3, c_4\}$, a substrate $\mathcal{V}$, a functionally important site index set $\mathcal{M}$, its residue set $\boldsymbol{s}_{\mathcal{M}}$ and the corresponding 3D coordinates $\boldsymbol{x}_{\mathcal{M}}$, generate an enzyme sequence $\boldsymbol{s}$ and $C_{\alpha}$ coordinates $\boldsymbol{x}$ of $N$ residues satisfying the substrate binding constraint $y$. 
Essentially, we aim to learn a generative model with probability $P(\boldsymbol{s}, \boldsymbol{x}, y|\boldsymbol{s}_{\mathcal{M}}, \boldsymbol{x}_{\mathcal{M}}, c, \mathcal{V})$. At inference time, we only use EC tag $c$, the functionally important sites $\boldsymbol{s}_M$ and $\boldsymbol{x}_M$ to generate an enzyme.

\subsection{Overall Model Architecture}
We propose a model named \model, as illustrated in Figure~\ref{Fig: model} (a), to simultaneously generate an enzyme sequence and its 3D backbone structure, constrained by the enzyme's small-molecule substrate. 
\model is a deep neural network consisting of an enzyme modeling module and a substrate representation module.
The enzyme modeling module is composed of $L_e$ stacked neighborhood attentive equivariant layers~({\layer}s).
Each \layer is composed of a global attention sub-layer using Transformer~\citep{vaswani2017attention} and a neighborhood equivariant sub-layer~(Figure~\ref{Fig: model} (b)) to incorporate information from the nearby residues based on $C_{\alpha}$ coordinates.
The substrate representation module consists of $L_s$ stacked neighborhood equivariant layers to pass messages inside the substrate $\mathcal{V}$, which will then provide binding constraint on the design of enzyme.

Suppose $\theta$ are \model parameters. We can formulate the joint probability as:
\begin{equation}
\footnotesize
\begin{split}
P(\boldsymbol{s}, \boldsymbol{x}, y&|\boldsymbol{s}_{\mathcal{M}}, \boldsymbol{x}_{\mathcal{M}}, c, \mathcal{V};\theta)=P(\boldsymbol{s}, \boldsymbol{x}|\boldsymbol{s}_{\mathcal{M}}, \boldsymbol{x}_{\mathcal{M}},c;\theta)\\
&\qquad\qquad\qquad\qquad\cdot P(y|\boldsymbol{s}_{\mathcal{M}}, \boldsymbol{x}_{\mathcal{M}},c, \mathcal{V};\theta) \\
P(\boldsymbol{s}, \boldsymbol{x}&|\boldsymbol{s}_{\mathcal{M}}, \boldsymbol{x}_{\mathcal{M}},c;\theta)=P(\boldsymbol{s}|\boldsymbol{s}_{\mathcal{M}}, \boldsymbol{x}_{\mathcal{M}},c;\theta)\\
&\qquad\qquad\qquad\quad\cdot P(\boldsymbol{x}|\boldsymbol{s}_{\mathcal{M}}, \boldsymbol{x}_{\mathcal{M}},c;\theta)\\
P(\boldsymbol{s}&|\boldsymbol{s}_{\mathcal{M}}, \boldsymbol{x}_{\mathcal{M}}, c;\theta)=\Pi_{i=1\& i\notin \mathcal{M}}^N P(\boldsymbol{s}_i|\boldsymbol{s}_{\mathcal{M}}, \boldsymbol{x}_{\mathcal{M}},c;\theta) \\ 
P(\boldsymbol{s}_i&|\boldsymbol{s}_{\mathcal{M}}, \boldsymbol{x}_{\mathcal{M}}, c;\theta)=\mathrm{Softmax}(W_{\mathcal{A}} \cdot \boldsymbol{h}_i^{L_e})\\
P( \boldsymbol{x}&|\boldsymbol{s}_{\mathcal{M}}, \boldsymbol{x}_{\mathcal{M}}, c;\theta) = \Pi_{i=1\&i\notin \mathcal{M}}^N P( \boldsymbol{x}_i|\boldsymbol{s}_{\mathcal{M}}, \boldsymbol{x}_{\mathcal{M}},c;\theta)\\ \boldsymbol{x}_i&|\boldsymbol{s}_{\mathcal{M}}, \boldsymbol{x}_{\mathcal{M}} \sim \mathcal{N}(\boldsymbol{x}_i^{L_e}; \lambda I) 
\end{split}
\end{equation}
where $W_{\mathcal{A}}$ is the embedding matrix for 20 common amino acids, $\boldsymbol{h}_i^{L_e}$ is the output embedding for $i^{th}$ residue at the last layer of enzyme modeling module and $\boldsymbol{x}_i^{L_e}$ is the corresponding output coordinate. $\mathcal{N}(\boldsymbol{x}_i^{L_e}; \lambda I)$ is the Gaussian distribution with mean $\boldsymbol{x}_i^{L_e}$ and covariance matrix $\lambda I$~($I$ is the identity matrix). $\lambda$ is a hyperparameter.
To find the optimal $\theta$, we maximize the conditional log likelihood~(i.e., minimizing the negative log likelihood):
\begin{equation}
\footnotesize
\begin{split}
\theta^* &= \mathop{\arg\min}\limits_{\theta}\mathcal{L}(\theta)=\mathop{\arg\min}\limits_{\theta} -\log P(\boldsymbol{s}, \boldsymbol{x}, y|\boldsymbol{s}_{\mathcal{M}}, \boldsymbol{x}_{\mathcal{M}},c, \mathcal{V};\theta)\\
&=\mathop{\arg\min}\limits_{\theta} \big\{-\sum\nolimits_{i=1\& i\notin \mathcal{M}}^N \log P(\boldsymbol{s}_i|\boldsymbol{s}_{\mathcal{M}}, \boldsymbol{x}_{\mathcal{M}},c;\theta)\\
&- \sum\nolimits_{i=1\&i\notin \mathcal{M}}^N \log P( \boldsymbol{x}_i|\boldsymbol{s}_{\mathcal{M}}, \boldsymbol{x}_{\mathcal{M}},c;\theta) \\
&-\log P(y|\boldsymbol{s}_{\mathcal{M}}, \boldsymbol{x}_{\mathcal{M}},c, \mathcal{V};\theta) \big\}
\end{split}
\end{equation}
For simplicity, we omit the number of enzyme samples in the dataset. The second log likelihood function can be further simplified as:
\begin{equation}
\small 
\log P( \boldsymbol{x}_i|\boldsymbol{s}_{\mathcal{M}}, \boldsymbol{x}_{\mathcal{M}},c;\theta) = -\frac{\lambda}{2} ||\boldsymbol{x}_i -\boldsymbol{x}_i^{L_e}||_2^2 + \mathrm{const}
\end{equation}
Therefore, the overall training objective is:
\begin{equation}
\footnotesize
\begin{split}
&\mathcal{L}(\theta)
=-\sum\nolimits_{i=1\& i\notin \mathcal{M}}^N \log P(\boldsymbol{s}_i|\boldsymbol{s}_{\mathcal{M}}, \boldsymbol{x}_{\mathcal{M}},c;\theta) \\&+ \frac{\lambda}{2}\sum\nolimits_{i=1\&i\notin \mathcal{M}}^N ||\boldsymbol{x}_i -\boldsymbol{x}_i^{L_e}||_2^2 - \log P(y|\boldsymbol{s}_{\mathcal{M}}, \boldsymbol{x}_{\mathcal{M}},c, \mathcal{V};\theta)
\end{split}
\label{equation_all}
\end{equation}


\subsection{\layer Global Attention Sub-Layer}
This sub-layer computes global contextual embeddings for all enzyme residues, which does not consider the closeness of residues in 3D space.
By allowing every residue to attend to all other residues across the whole sequence, we facilitate information flow through the entire enzyme sequence. 

We adopt the Transformer layer~\citep{vaswani2017attention} to compute global contextual embeddings. Specifically, each transformer layer is composed of one multi-head self-attention sub-layer~(MHA) and one fully connected feed-forward network~(FFN). A residue connection and a layer normalization are employed after each of the two sub-layers.
The calculation of the global sequence attention can be formulated as follows:
\begin{equation}
\small 
\begin{split}
\boldsymbol{h}_i^{l+0.5}&=\mathrm{LayerNorm} \left(\mathrm{FFN}(\Tilde{\boldsymbol{h}}_i^{l+0.5})+\Tilde{\boldsymbol{h}}_i^{l+0.5} \right), \\ \Tilde{\boldsymbol{h}}_i^{l+0.5}&=\mathrm{LayerNorm}\left(\mathrm{MHA}(\boldsymbol{h}_i^l, \boldsymbol{H}_e^l)+\boldsymbol{h}_i^l \right)
\end{split}
\end{equation}
where $\boldsymbol{h}_i^{l}$ is the $i^{th}$ residue input representation at $l^{\mathrm{th}}$ layer and $\boldsymbol{H}_e^l=[\boldsymbol{h}_1^l, \boldsymbol{h}_2^l, ..., \boldsymbol{h}_N^l]^T$. 
The input residue embeddings for the first layer are either taken from an embedding lookup table for functionally important sites, or initialized with a special [mask] token embedding for other residues. The residue embedding will then be enhanced by the enzyme family embedding Emb($c$) corresponding to the desired enzyme family $c=\{c_1, c_2, c_3, c_4\}$:
\begin{equation} 
\small 
\begin{split}
\boldsymbol{h}_i^0 &= \hat{\boldsymbol{h}}_i^0 + \mathrm{Emb}(c_1) + \mathrm{Emb}(c_2) + \mathrm{Emb}(c_3) + \mathrm{Emb}(c_4)\\
\hat{\boldsymbol{h}}_i^0 &= \left\{
\begin{aligned}
W_{\mathcal{A}}^T \boldsymbol{s}_i &,  & {\text{$i \in \mathcal{M}$,}} \\
\mathrm{Emb([mask]}) &, & {\text{otherwise}}
\end{aligned}
\right.
\end{split}
\end{equation}
Similar to token-level language tags employed in multilingual models~\cite{chi2020cross,song2021switch}, we also utilize the enzyme family tags at the residue level.


\subsection{\layer Neighborhood Equivariant Sub-Layer}
Properly modeling the interactions of a given residue and its nearest neighboring residues in the 3D space can lead to improved residue representations. We intend to model such impact with a carefully designed subnetwork while keeping the equivariance under 3D translation and rotation. To this end, we propose the neighborhood equivariant sub-layer. 
This sub-layer includes three components: neighborhood message update, neighborhood coordinate update and neighborhood node feature update~(Figure~\ref{Fig: model} (b)).
Updating residue representations and coordinates in 3D space with only nearest neighbors enables more efficient and economic message passing compared to prior approaches which compute messages on the complete pairwise residue graph. 

\textbf{Neighborhood message update.}
We first compute distances between residues using $C_{\alpha}$ coordinates, and select $K$ nearest residues~(Figure~\ref{Fig: model} (b) green region). 
We compute the messages between $i^{th}$ residue and its $K$-nearest neighbors~(denoted as Neighbor(i)) as follows:
\begin{equation}
\footnotesize 
\begin{split}
\boldsymbol{m}_{ik}^{l+0.5} &= \mathrm{SiLU}(\mathrm{FFN}([(\boldsymbol{h}_i^{l+0.5}; \boldsymbol{h}_k^{l+0.5}; ||\boldsymbol{x}_i^{l}-\boldsymbol{x}^{l}_{k}||_2])) \\
w_{ik}^{l+0.5} &= \frac{\exp (W_a^l\boldsymbol{m}_{ik}^{l+0.5}+b_a^l)}{\sum_{k'\in \mathrm{Neighbor(i)}} \exp (W_a^l\boldsymbol{m}_{ik'}^{l+0.5}+b_a^l)}\\
\boldsymbol{m}_{ik}^{l+1} &= w_{ik}^{l+0.5} * \boldsymbol{m}_{ik}^{l+0.5} \\
\end{split}
\end{equation}
where FFN is a two-layer fully connected feed-forward network with SiLU activation function after its first layer.
[;] is concatenation operator and $||\boldsymbol{x}_i^{l}-\boldsymbol{x}^{l}_{k}||_2$ is the Euclidean distance between $i^{th}$ and $k^{th}$ residue coordinates at $l^{th}$ layer. 
$W_*$ and $b_*$ are trainable parameters.

\textbf{Neighborhood coordinate update.} We update the $C_{\alpha}$ coordinate of $i^{th}$ residue as a $K$-nearest neighbor vector field in a radial direction (Figure~\ref{Fig: model} (b) blue region). 
The $C_{\alpha}$ coordinate $\boldsymbol{x}^{l}_i$ at $l^{th}$ layer of $i^{th}$ residue is updated with the weighted sum of all relative differences $(\boldsymbol{x}_i^{l}-\boldsymbol{x}_k^{l})_{\forall{k\in \mathrm{Neighbor(i)}}}$:
\begin{equation}
\small 
\begin{split}
\boldsymbol{x}^{l+1}_i=\boldsymbol{x}^{l}_i+\sum\nolimits_{k\in \mathrm{Neighbor(i)}}(\boldsymbol{x}^{l}_i-\boldsymbol{x}^{l}_k)\cdot \mathrm{FFN}(\boldsymbol{m}_{ik}^{l+1})
\end{split}
\label{eq:neighbor_coordinate}
\end{equation}

The input residue $C_{\alpha}$ coordinates are either the given coordinates for functionally important sites, or randomly initialized as 3D points on the spherical surface centered at its preceding residue, considering the Euclidean distances between neighboring $C_\alpha$ pairs are almost the same (around $r=3.75${\small \AA}):
\[
\small 
\boldsymbol{x}_i^0 = \left\{
\begin{aligned}
&\boldsymbol{x}_i  & \text{$i \in \mathcal{M}$}  \\
&\boldsymbol{x}_{i-1}^0+r\cdot[\sin{\omega_1}\cos{\omega_2}, \sin{\omega_1}\sin{\omega_2}, \cos{\omega_1}]^T & \text{$i \notin \mathcal{M}$}
\end{aligned}
\right.
\]
where $\omega_1\sim \mathrm{Uniform}(0, \pi)$ is the angle to Z-axis and $\omega_2\sim \mathrm{Uniform}(0,2\pi)$ is the angle to X-axis in polar coordinate system.

\textbf{Neighborhood node feature update.} We update the $i^{th}$ residue feature by gathering information from its $K$-nearest neighbors using a gating mechanism~(Figure~\ref{Fig: model} (b) red region):
\begin{equation}
\footnotesize 
\begin{split}
\boldsymbol{g}_{i}^{l+1} &= \sum\nolimits_{k\in \mathrm{Neighbor(i)}} \boldsymbol{m}_{ik}^{l+1} \\
\boldsymbol{h}_{i}^{l+1} &= \boldsymbol{h}_{i}^{l+0.5} + \sigma(\mathrm{FFN}(\boldsymbol{g}_{i}^{l+1})) \odot \boldsymbol{g}_{i}^{l+1}
\end{split}
\label{eq:neighbor_node}
\end{equation}
where FFN is a two-layer fully connected feed-forward network with ReLU activation function after its first layer. $\sigma$ denotes the sigmoid activation function.

We stack $L_e$ layers of \layer to model enzymes. 
The output embedding $\boldsymbol{h}^{L_{e}}_i$ and coordinate $\boldsymbol{x}^{L_{e}}_i$ for the $i^{th}$ residue are both from the last layer.
Then the output probability of amino acid type for $i^{th}$ residue is calculated as:
\begin{equation}
\small 
P(s_i=a|\boldsymbol{s}_{\mathcal{M}}, \boldsymbol{x}_{\mathcal{M}},c)=\frac{\exp(h_{i, a}^o)}{\sum\nolimits_{a'=1}^{20} \exp(h_{i, a'}^o)}, \quad \boldsymbol{h}_i^o=W_{\mathcal{A}}\cdot \boldsymbol{h}_i^{L_e}
\end{equation}

\begin{figure}
  \centering
  \includegraphics[width=6.0cm]{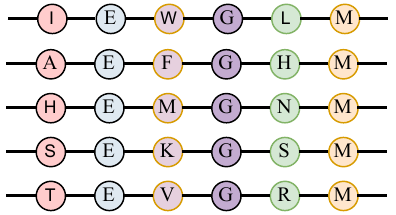}
  \vspace{-0.6em}
  \caption{Discoverying functionally important sites. Each row is a protein sequence in a same enzyme family (BRENDA fourth-level enzyme classification tree category). We use ClustalW2 to perform multiple sequence alignment and select common residuals above the identity threshold $\tau$.  According to the aligned sequences, E, G and M are common in all the sequences therefore these are selected as important sites. In experiment, $\tau=30\%$.}
  \label{Fig: important_site_example}
\end{figure}

\subsection{Substrate Representation Module}
To facilitate effective and efficient message passing inside the substrate, we stack $L_s$ neighborhood equivariant layers to learn the substrate representations. With the substrate atom embedding $\boldsymbol{h}_j^l$
and the corresponding coordinate $\boldsymbol{x}_j^l$ at the $l^{th}$ layer, we perform the substrate message passing as follows:
\begin{equation}
\small 
\begin{split}
\boldsymbol{m}_{jk}^{l+1} &= \phi_m(\boldsymbol{h}_j^{l},\boldsymbol{h}_k^{l}, ||\boldsymbol{x}_j^{l}-\boldsymbol{x}^{l}_{k}||_2) \\
\boldsymbol{h}_{j}^{l+1} &= \boldsymbol{h}_{j}^{l} + \phi_n(\boldsymbol{h}_{j}^{l}, \sum\nolimits_{k\in \mathrm{Neighbor(j)}} \boldsymbol{m}_{jk}^{l+1})
\end{split}
\end{equation}
where $\phi_m$ and $\phi_n$ respectively denotes neighborhood message update~Eq.\eqref{eq:neighbor_coordinate} and neighborhood node feature update~Eq.\eqref{eq:neighbor_node} with separate parameters. Following \citet{guan20223d}, we initialize the node feature as five chemical features $\boldsymbol{\hat{h}}_j^0\in\mathbb{R}^{5}$ and $\boldsymbol{h}_j^0=W_s\cdot \boldsymbol{\hat{h}}_j^0$ where $W_s\in \mathbb{R}^{d\times 5}$ is a mapping matrix and $d$ is the representation dimensionality. Since we have the real 3D structure of the substrate, we do not need to update its atom coordinates.

Then based on the learned enzyme and substrate representations, we can predict whether the enzyme can bind with the given substrate as follows:
\begin{equation}
\small 
P(y|\boldsymbol{s}_{\mathcal{M}}, \boldsymbol{x}_{\mathcal{M}},c, \mathcal{V};\theta)=\mathrm{Softmax}(W_b\cdot[f(\boldsymbol{H}^{L_e}_e);f(\boldsymbol{H}^{L_s}_s)])
\end{equation}
where $W_b\in \mathbb{R}^{2\times 2d}$ is a mapping matrix and $f$ denotes sum pooling operation. $\boldsymbol{H}^{L_e}_e=[\boldsymbol{h}^{L_e}_1,...,\boldsymbol{h}^{L_e}_N]^T$ and $\boldsymbol{H}^{L_s}_s=[\boldsymbol{h}^{L_s}_1,...,\boldsymbol{h}^{L_s}_m]^T$ denote the output embeddings of enzyme and substrate at the last layer of enzyme modeling module and substrate representation module, respectively.

\subsection{Functionally Important Enzyme Site Discovery}
 Functional sites are often evolutionarily conserved in a given enzyme family~\cite{tristem2000molecular,bickel2002finding}. Following previous methods~\cite{armon2001consurf,panchenko2004prediction,chakrabarti2007analysis,bray2009sitesidentify}, we leverage multiple sequence alignment~(MSA) to mine evolutionarily conserved patterns, which are known as functionally or structurally important sites. 
Designing enzymes conditioning on these conserved patterns can facilitate the generation of functional enzymes.
We employ the ClustalW2 method~\citep{anderson2011suitemsa} to perform multiple sequence alignment for all proteins in each enzyme family corresponding to BRENDA fourth-level EC tree category. 
We select the residues that occur at the same site in more than $\tau$ sequences. These residues are designated as functionally important sites. 
In practice, we set $\tau$ to be $30\%$.
Figure~\ref{Fig: important_site_example} illustrate the procedure to select functionally important sites.



\begin{table}[!t]
\small
\begin{center}
\begin{tabular}{lccc}
\midrule
Dataset  & Training  & Validation  & Test \\
\midrule
Third-Level Category & 256 & 30 & 30 \\
Fourth-Level Category & 3135 & 428 & 323 \\
Size & 98974 & 1500 & 1500  \\
Average Seq. Length & 327 & 331 & 328 \\
\bottomrule
\end{tabular}
\end{center}
\caption{\dataset statistics. The first two rows are third and fourth-level categories in BRENDA enzyme classification tree. Size indicates the number of protein entries in PDB under the categories. }
\label{Tab: data_statistics}
\end{table}

\section{Experiments}
\label{experiments}

\begin{table*}[!t]
\scriptsize
\centering
\setlength{\tabcolsep}{1.8mm}
\begin{tabular}{lcccccccccccccccc}
\midrule
Enzyme Family & 1.1.1 & 1.11.1 & 1.14.13 & 1.14.14 & 1.2.1 & 2.1.1 & 2.3.1 & 2.4.1 & 2.4.2 & 2.5.1 & 2.6.1 & 2.7.1 & 2.7.10 & 2.7.11 & 2.7.4   \\
\cmidrule(r){1-16}
PROTSEED & 0.54&0.67&0.24&0.39&0.57&\textbf{0.83}&0.52&0.29&0.75&0.58&0.45&0.77&0.88&0.81&0.78 &   \\
RFDiffusion+IF & 0.45&0.94&\textbf{0.54}&0.39&0.47&0.43&0.48&\textbf{0.39}&0.52&0.46&0.53&0.50&0.51&0.60&0.55 &  \\
ESM2+EGNN &0.58&0.90&0.35&0.35&0.63&0.79&0.53&0.32&0.80&0.59&0.51&0.76&0.88&0.88&0.77 &  \\
\rowcolor{myblue}
\model&\textbf{0.64}&\textbf{0.98}&0.38&\textbf{0.42}&\textbf{0.72}&0.80&\textbf{0.61}&0.38&\textbf{0.86}&\textbf{0.66}&\textbf{0.53}&\textbf{0.76}&\textbf{0.92}&\textbf{0.93}&\textbf{0.80}   \\
\hdashline
\rowcolor{myblue}
\model-1.5 &\textbf{0.66}&\textbf{0.99}&\textbf{0.50}&\textbf{0.74}&0.70&\textbf{0.85}&0.60&\textbf{0.70}&\textbf{0.86}&\textbf{0.66}&\textbf{0.53}&\textbf{0.77}&0.91&\textbf{0.93}&\textbf{0.81}   \\
\midrule
Enzyme Family & 2.7.7 & 3.1.1 & 3.1.3 & 3.1.4 & 3.2.2 & 3.4.19 & 3.4.21 & 3.5.1 & 3.5.2 & 3.6.1 & 3.6.4 & 3.6.5 & 4.1.1 & 4.2.1 & 4.6.1 & Avg  \\
\midrule
PROTSEED & 0.69&0.70&\textbf{0.90}&0.84&0.48&0.29&0.69&0.31&0.10&0.50&0.57&0.37&0.84&0.83&0.42&0.59  \\
RFDiffusion+IF & 0.53&0.33&0.61&0.62&\textbf{0.49}&\textbf{0.62}&0.45&\textbf{0.47}&\textbf{0.44}&0.55&0.63&\textbf{0.59}&0.59&0.84&0.45&0.53  \\
ESM2+EGNN &0.70&0.71&0.78&0.82&0.43&0.22&0.56&0.35&0.11&0.61&0.73&0.37&\textbf{0.81}&0.89&0.54&0.61 \\
\rowcolor{myblue}
\model&\textbf{0.79}&\textbf{0.76}&0.62&\textbf{0.88}&0.47&0.26&\textbf{0.73}&0.40&0.14&\textbf{0.66}&\textbf{0.78}&0.40&0.80&\textbf{0.93}&\textbf{0.57}&\textbf{0.65} \\
\hdashline
\rowcolor{myblue}
\model-1.5&\textbf{0.80}&0.75&0.56&0.86&0.45&0.25&0.68&\textbf{0.58}&0.20&\textbf{0.67}&0.77&0.40&0.45&\textbf{0.93}&0.55&\textbf{0.67} \\
\bottomrule
\end{tabular}
\vspace{-0.6em}
\caption{Enzyme-substrate interaction~(ESP) scores ($\uparrow$) of generated enzymes for 30 testing BRENDA EC categories averaged over 50 cases in each category. IF denotes the inverse folding model ProteinMPNN. A ESP score of 0.6 or higher indicates a positive substrate binding. The last column (Avg) is the  average across 30 categories. Notice that our \model outperforms all previous methods in ESP by a big margin.}
\label{Tab: ESP_all}
\end{table*}

\begin{table*}[!t]
\scriptsize
\centering
\setlength{\tabcolsep}{1.60mm}
\begin{tabular}{lcccccccccccccccc}
\midrule
Enzyme Family & 1.1.1 & 1.11.1 & 1.14.13 & 1.14.14 & 1.2.1 & 2.1.1 & 2.3.1 & 2.4.1 & 2.4.2 & 2.5.1 & 2.6.1 & 2.7.1 & 2.7.10 & 2.7.11 & 2.7.4   \\
\cmidrule(r){1-16}
PROTSEED &  -6.61&-3.93&-4.27&-10.22&-6.71&-8.84&-9.58&-7.43&-10.01&-7.44&-5.46&-8.07&-9.68&-10.24&-11.68  \\
RFDiffusion+IF & -7.11&-4.43&-4.70&\textbf{-10.74}&\textbf{-7.21}&-9.61&\textbf{-10.04}&-7.93&-10.64&-7.84&-6.19&-8.55&-10.60&-10.44&\textbf{-12.18}    \\
ESM2+EGNN &-6.66&-4.47&-4.81&-10.73&-7.02&-9.57&-9.98&-8.61&-10.90&-7.95&-6.43&-8.79&-10.23&-10.75&-11.31 \\
\rowcolor{myblue}
\model & \textbf{-8.44}&\textbf{-4.58}&\textbf{-5.10}&-10.34&-6.95&\textbf{-10.05}&-9.89&\textbf{-9.65}&\textbf{-11.91}&\textbf{-9.98}&\textbf{-8.05}&\textbf{-10.50}&\textbf{-11.65}&\textbf{-12.51}&-11.24  \\
\hdashline
\rowcolor{myblue}
\model-1.5 & -7.63&\textbf{-5.26}&\textbf{-5.87}&-10.33&\textbf{-7.87}&\textbf{-11.11}&\textbf{-10.61}&\textbf{-10.91}&-11.84&-9.19&-7.46&-9.71&\textbf{-11.65}&-12.10&\textbf{-12.57}  \\
\midrule
Enzyme Family & 2.7.7 & 3.1.1 & 3.1.3 & 3.1.4 & 3.2.2 & 3.4.19 & 3.4.21 & 3.5.1 & 3.5.2 & 3.6.1 & 3.6.4 & 3.6.5 & 4.1.1 & 4.2.1 & 4.6.1 & Avg  \\
\midrule
PROTSEED & -8.00&-6.01&-7.20&-9.16&-9.53&-8.79&-8.67&-5.19&-5.44&-8.57&-10.11&-9.37&-9.74&-4.38&-9.11&-7.94  \\
RFDiffusion+IF & -8.50&-6.51&-7.70&-11.65&-10.08&-9.29&-9.03&-5.54&-5.94&-9.07&-11.12&-9.87&-11.24&-4.88&-9.68&-8.57  \\
ESM2+EGNN &-8.47&-5.81&-7.20&-11.45&-10.14&-8.91&-9.25&-5.59&-5.42&-8.23&-10.69&-11.15&\textbf{-11.34}&-5.01&-9.76&-8.52 \\
\rowcolor{myblue}
\model & \textbf{-8.86}&\textbf{-7.01}&\textbf{-8.89}&\textbf{-11.90}&\textbf{-10.50}&\textbf{-10.60}&\textbf{-10.49}&\textbf{-6.37}&\textbf{-6.84}&\textbf{-9.23}&\textbf{-13.10}&\textbf{-11.35}&-11.03&\textbf{-5.51}&\textbf{-10.64}&\textbf{-9.44} \\
\hdashline
\rowcolor{myblue}
\model-1.5 & \textbf{-9.55}&-6.67&-8.64&-12.01&\textbf{-11.41}&-10.14&\textbf{-10.74}&\textbf{-6.72}&-6.38&-9.10&-12.15&\textbf{-12.39}&-10.38&\textbf{-6.04}&\textbf{-10.88}&\textbf{-9.58} \\
\bottomrule
\end{tabular}
\vspace{-0.6em}
\caption{Enzyme-substrate binding affinity~($\downarrow$)~(kcal/mol) for 30 testing BRENDA EC categories, evaluated by a docking software Gnina. IF denotes the inverse folding model ProteinMPNN. The last column (Avg) is the  average across 30 categories. Notice that our \model outperforms all previous methods in binding affinity by a big margin.}
\label{Tab: docking_all}
\end{table*}

\begin{table*}[!t]
\scriptsize
\centering
\setlength{\tabcolsep}{1.65mm}
\begin{tabular}{lcccccccccccccccc}
\midrule
Enzyme Family & 1.1.1 & 1.11.1 & 1.14.13 & 1.14.14 & 1.2.1 & 2.1.1 & 2.3.1 & 2.4.1 & 2.4.2 & 2.5.1 & 2.6.1 & 2.7.1 & 2.7.10 & 2.7.11 & 2.7.4   \\
\cmidrule(r){1-16}
PROTSEED &  77.10&81.80&71.19&74.24&78.67&77.40&74.54&75.18&77.11&74.79&75.55&75.90&81.05&74.05&76.50  \\
RFDiffusion+IF & 82.47&89.63&81.12&89.32&82.04&82.49&85.14&85.61&81.13&86.25&81.60&\textbf{87.51}&86.75&85.91&88.30  \\
ESM2+EGNN &90.67&93.11&90.93&90.30&87.67&79.40&84.78&84.80&84.56&\textbf{90.21}&87.47&83.52&\textbf{88.92}&85.59&90.09 \\
\rowcolor{myblue}
\model & \textbf{91.86}&\textbf{94.40}&\textbf{93.02}&\textbf{92.70}&\textbf{91.99}&\textbf{83.47}&\textbf{87.71}&\textbf{92.81}&\textbf{87.02}&89.69&\textbf{89.20}&85.19&87.55&\textbf{87.64}&\textbf{91.81}  \\
\hdashline
\rowcolor{myblue}
\model-1.5 & 91.63&\textbf{95.19}&92.91&92.60&89.89&\textbf{84.95}&86.79&91.12&\textbf{87.78}&89.21&88.85&85.12&86.98&\textbf{88.18}&\textbf{91.85}  \\
\midrule
Enzyme Family & 2.7.7 & 3.1.1 & 3.1.3 & 3.1.4 & 3.2.2 & 3.4.19 & 3.4.21 & 3.5.1 & 3.5.2 & 3.6.1 & 3.6.4 & 3.6.5 & 4.1.1 & 4.2.1 & 4.6.1 & Avg  \\
\midrule
PROTSEED & 78.00&76.29&77.89&75.75&78.76&73.56&82.40&76.70&75.90&75.16&74.62&83.46&76.36&78.87&83.31&77.07 \\
RFDiffusion+IF & 81.25&80.01&81.59&81.22&\textbf{92.04}&\textbf{89.72}&77.20&84.05&85.47&71.35&\textbf{82.87}&\textbf{84.49}&81.31&79.02&76.11&83.43  \\
ESM2+EGNN &81.80&87.27&\textbf{87.05}&85.50&72.23&71.31&82.62&83.48&88.69&84.96&73.34&80.77&\textbf{87.72}&89.70&85.48&85.13\\
\rowcolor{myblue}
\model & \textbf{83.75}&\textbf{89.79}&85.40&\textbf{89.68}&74.44&77.14&\textbf{89.11}&\textbf{86.70}&\textbf{89.80}&\textbf{85.98}&76.31&84.32&85.71&\textbf{91.88}&\textbf{87.55}&\textbf{87.45}\\
\hdashline
\rowcolor{myblue}
\model-1.5 & 83.46&\textbf{90.47}&78.29&88.83&\textbf{74.84}&76.28&84.29&\textbf{87.36}&\textbf{90.58}&\textbf{87.63}&\textbf{77.12}&\textbf{86.39}&60.65&\textbf{92.14}&83.10&86.15\\
\bottomrule
\end{tabular}
\vspace{-0.6em}
\caption{pLDDT~($\uparrow$) predicted by AlphaFold2. IF denotes the inverse folding model ProteinMPNN. The last column (Avg) is the  average across 30 categories. Notice that our \model outperforms all previous methods in pLDDT. It shows \model's enzyme can fold more stably. }
\label{Tab: plddt_all}
\end{table*}

\subsection{\dataset Construction}
We aim to evaluate the effectiveness of our \model across a diverse range of enzymes. To achieve this, we gather all enzymes from BRENDA~\cite{schomburg2002brenda}, yielding a total of $101,974$ PDB entries. These entries are classified into $3,157$ fourth-level enzyme categories based on the EC Tree (Appendix Figure~\ref{Fig: appendix_ec_tree}).  It is noteworthy that there exist a total of $8,422$ fourth-level categories; however, our focus lies specifically on those with experimentally confirmed structures. This focus results in our dataset comprising $3,157$ fourth-level categories. We then conduct MSAs for each fourth-level category to identify functionally important sites, using an MSA threshold ($\tau$) of 30\%. Given the relatively small size of fourth-level categories, we aggregate data belonging to the same third-level category in the EC Tree, resulting in $256$ third-level categories.  
Then we select 30 third-level categories for validation and testing, respectively including 428 and 323 fourth-level categories.
For each of the 30 third-level categories, we we randomly split $100$ PDB entries with 50 for validation and 50 for testing, while the remaining entries are utilized for training.
To prevent data leakage, we cluster the PDB entries with a sequence identity threshold of $50\%$. We ensure that no PDB entries in the validation or test sets belong to the same cluster as those in the training set. 
We further collect $23,711$ experimentally confirmed $<$enzyme, substrate$>$ pairs from \citet{kroll2023general}, ensuring enzymes without a corresponding substrate have a randomly sampled negative one during the training process. 
We also make sure that all entries in the test set have a corresponding substrate to ensure function evaluation. 
Detailed data statistics are provided in Table~\ref{Tab: data_statistics} and Appendix~\ref{appendix:data_statistics}.

\subsection{Experimental Setup}
\textbf{Implementation Details.} 
We use 33 {\layer}s to model enzymes and 3 neighborhood equivariant layers for substrate representation module. The node feature dimensionality is set to 1280.
As suggested by \citet{ying2018hierarchical}, the number of graph layers typically ranges from 2 to 6. To prevent over-fitting and facilitate efficiency, we utilize 3 neighborhood equivariant sub-layers for enzyme modeling. The resulting \model consists of a modified NAEL with every 11 global attention sub-layers followed by one neighborhood equivariant sub-layer.
The parameters of the global attention sub-layer are initialized with the released 650M ESM-2~\cite{lin2022language} parameters.   The total number of parameter is 714 million. The variance and neighbor hyperparameters $\lambda/2$ and $K$ are set to $1.0$ and $30$. The model undergoes training for $1,000,000$ steps using 8 NVIDIA RTX A$6000$ GPUs. The model is trained with only the sequence generation and position prediction losses for the first $200,000$ steps, and then continues training on the sequence generation, position prediction and enzyme-substrate interaction prediction losses for $800,000$ steps.
The batch size and learning rate are set to $8192$ residues and $3$e-$4$ respectively.
Sequences are decoded using the greedy decoding strategy. We incorporate four-level enzyme tags in the EC tree. For example, an enzyme in the category 1.1.1.1 (alcohol dehydrogenase) should have four tags: 1 (oxidoreductases), 1.1 (acting on the CH-OH group of donors), 1.1.1 (with NAD+ or NADP+ as acceptor), and 1.1.1.1 (alcohol dehydrogenase). To further enhance model performance, we first pretrain the model using a masked language modeling (MLM) objective on both sequence and backbone structure, randomly masking 20\% of residues and $C_{\alpha}$ coordinates over 800,000 steps. We then follow the standard \model training procedure. We refer to this variant as \textbf{\model-1.5}.

\textbf{Baseline Models.}
We compare the proposed \model against the following representative baselines:
(1) \textbf{ESM2+EGNN} employs ESM2~\cite{lin2022language} as the sequence encoder and EGNN~\cite{satorras2021n} as the backbone structure predictor. The node features in EGNN are initialized with the output from ESM2. Both ESM2 and EGNN share the same layer configurations as our \model.
(2) \textbf{PROTSEED}~\citep{shi2022protein} co-designs protein sequences and backbone structures based on secondary structure and binary contact maps.
(3) \textbf{RFDiffusion}+\textbf{ProteinMPNN
} first applies RFDiffusion~\cite{watson2023novo} to design an enzyme structure based on the given functionally important sites and then uses ProteinMPNN~\cite{dauparas2022robust} to design a sequence based on the generated structure.
To ensure a fair and reliable comparison, we train ESM2+EGNN, PROTSEED and ProteinMPNN on our \dataset using their official codes and implementations. Since RFDiffusion does not provide a training script, we encountered challenges in training their code and reproducing their results. Consequently, we directly apply their released model to design the structures.



\textbf{Function Evaluation.}
We evaluate the enzyme-substrate interaction function by the following metrics: (1) Following previous work~\cite{li2013crystal,vidal2022integration}, we use \textbf{enzyme-substrate binding affinity}, calculated by Gnina~\citep{mcnutt2021gnina} to quantify the strength of interactions between the designed enzymes and their substrates. A lower binding affinity indicates stronger binding. (2) We also use \textbf{ESP score}, developed by \citet{kroll2023general} to assess the ability of the designed enzymes to bind their corresponding substrates. In their paper, ESP model predicts enzyme-substrate interaction with 91\% accuracy across multiple benchmarks.
The scale of ESP score is 0-1, and a higher ESP score indicates a stronger enzyme-substrate interaction. Additionally, to evaluate if the designed enzymes are well-folded, we compute (3) \textbf{pLDDT} using AlphaFold2.

\begin{figure*}
\begin{minipage}[t]{0.25\linewidth}
\centering
\includegraphics[width=4.1cm]{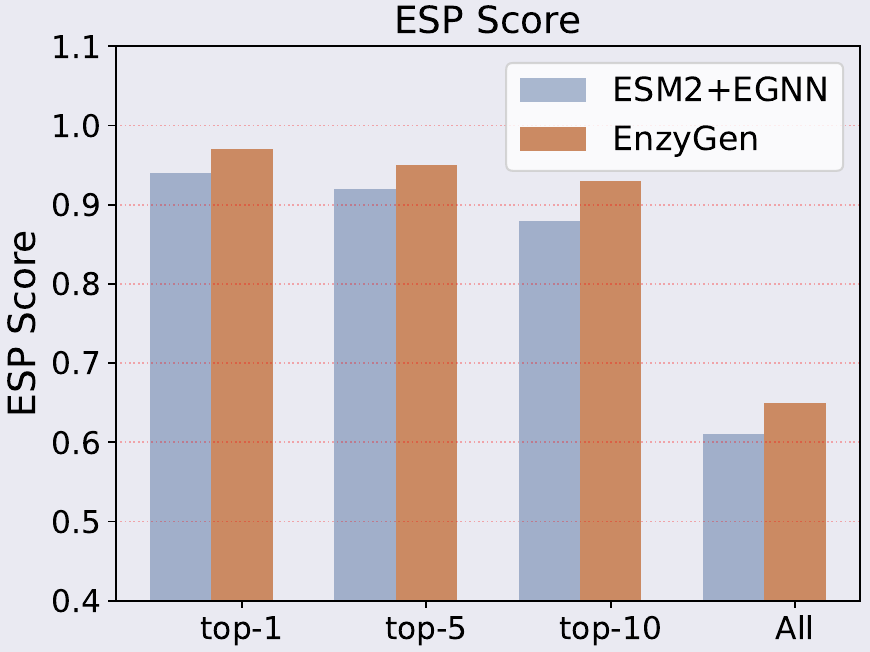}
\centerline{\small{(a) ESP score ($\uparrow$)}}
\end{minipage}%
\begin{minipage}[t]{0.25\linewidth}
\centering
\includegraphics[width=4.1cm]{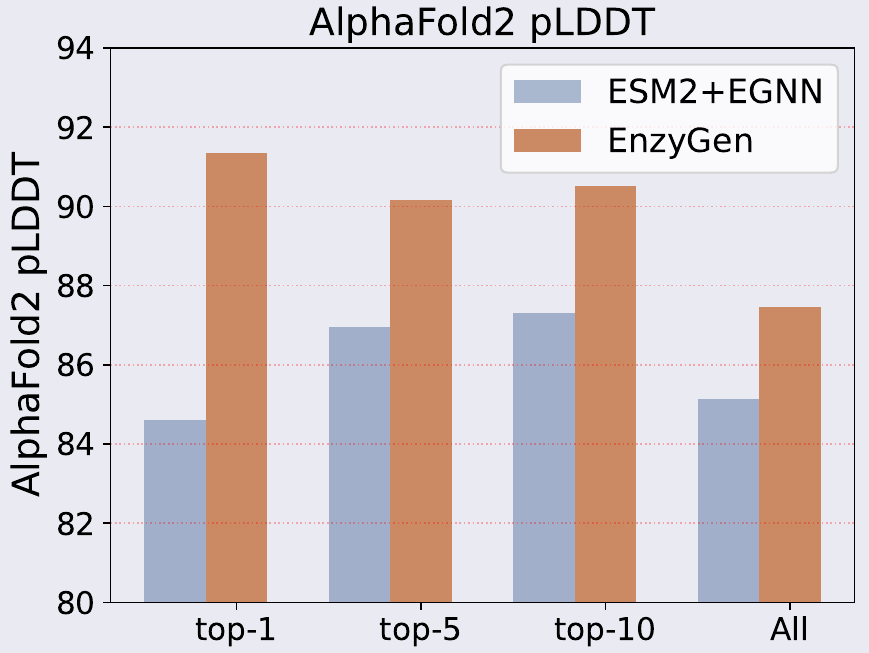}
\centerline{\small{(b) pLDDT ($\uparrow$)}}
\end{minipage}%
\begin{minipage}[t]{0.25\linewidth}
\centering
\includegraphics[width=4.15cm]{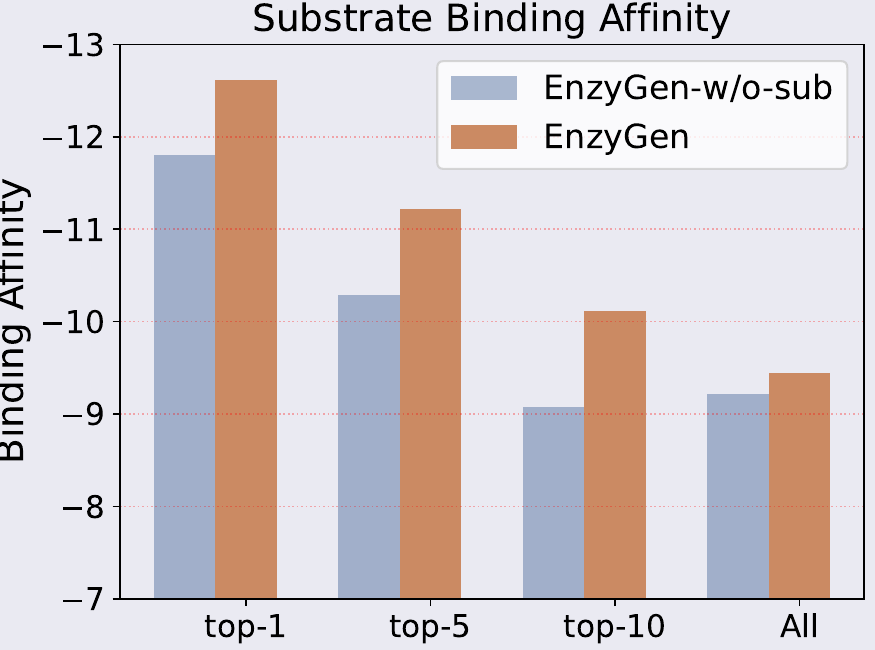}
\centerline{\small{(c) Substrate binding affinity ($\downarrow$)}}
\end{minipage}%
\begin{minipage}[t]{0.25\linewidth}
\centering
\includegraphics[width=4.05cm]{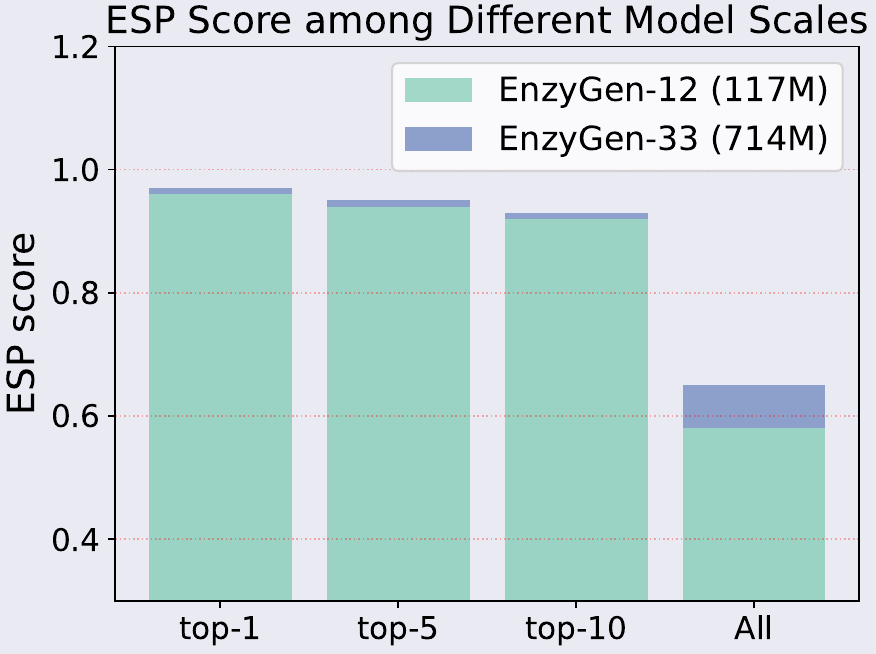}
\centerline{\small{(d) ESP score ($\uparrow$)}}
\end{minipage}
	\caption{(a) Ablation study comparing \model against ESM2+EGNN on ESP score. (b) Ablation study comparing \model against ESM2+EGNN on AlphaFold2 pLDDT. (c) Ablation study comparing \model against the model removing enzyme-substrate interaction constraint~(\model-w/o-sub). (d) Ablation study on different model scales. } 
 \label{Fig: ablation_interleaving}
\end{figure*}

\begin{figure}
\begin{minipage}[t]{0.5\linewidth}
\centering
\includegraphics[width=3.65cm]{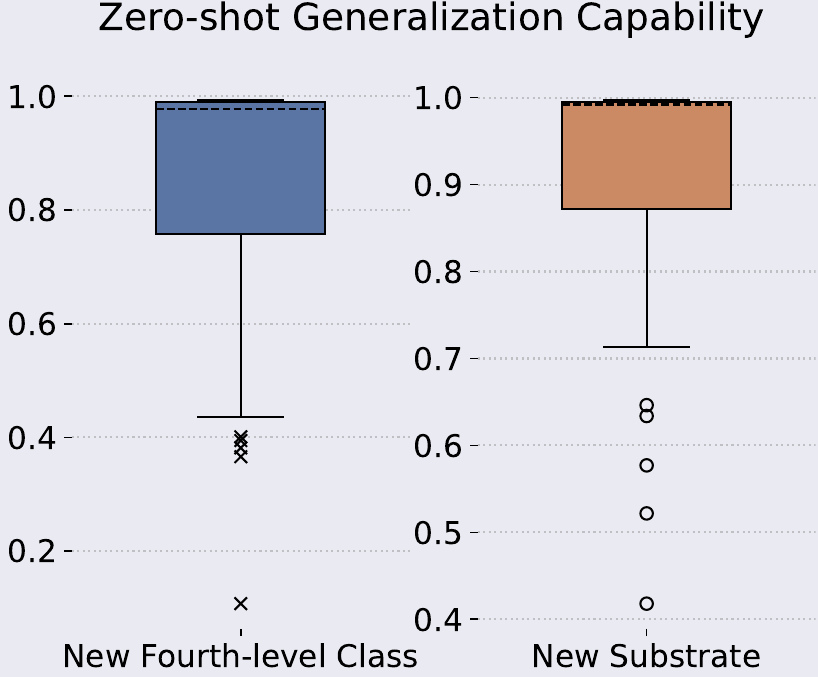}
\centerline{\small{(a) Zero-shot generalization}}
\end{minipage}%
\begin{minipage}[t]{0.5\linewidth}
\centering
\includegraphics[width=4.2cm]{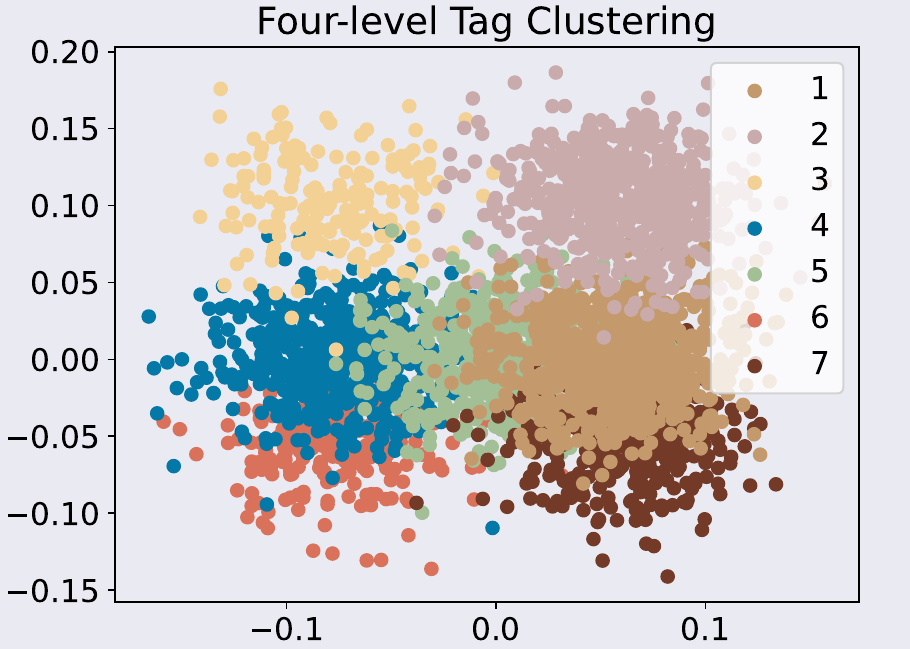}
\centerline{\small{(b) Fourth-level category clustering}}
\end{minipage}
	\caption{(a) ESP scores of designed enzymes from new fourth-level classes or with new substrates. Dash line denotes median. (b) Fourth-level category embedding clustering.} 
 \label{Fig: analysis}
\end{figure}

\subsection{Main Results}
The ESP scores, substrate binding affinities, and pLDDT results are shown in Tables~\ref{Tab: ESP_all}, \ref{Tab: docking_all} and \ref{Tab: plddt_all}. For clarity, we average the performance of fourth-level categories under the same third-level one, with detailed results for 323 families in Appendix~\ref{appendix_299_family_results}. The results for top-1, top-5 and top-10 candidate are also provided in Appendix~\ref{Appendix_different_candidate}.

\begin{table*}[!t]
\scriptsize
\centering
\setlength{\tabcolsep}{1.8mm}
\begin{tabular}{lcccccccccccccccc}
\midrule
Enzyme Family & 1.1.1 & 1.11.1 & 1.14.13 & 1.14.14 & 1.2.1 & 2.1.1 & 2.3.1 & 2.4.1 & 2.4.2 & 2.5.1 & 2.6.1 & 2.7.1 & 2.7.10 & 2.7.11 & 2.7.4   \\
\midrule
\model-finetune &\textbf{0.68}&0.95&0.38&0.37&0.68&\textbf{0.82}&0.57&0.38&0.86&\textbf{0.69}&\textbf{0.54}&\textbf{0.80}&\textbf{0.93}&\textbf{0.94}&\textbf{0.82}  \\
\rowcolor{myblue}
\model&0.64&\textbf{0.98}&0.38&\textbf{0.42}&\textbf{0.72}&0.80&\textbf{0.61}&0.38&0.86&0.66&0.53&0.76&0.92&0.93&0.80   \\
\midrule
Enzyme Family & 2.7.7 & 3.1.1 & 3.1.3 & 3.1.4 & 3.2.2 & 3.4.19 & 3.4.21 & 3.5.1 & 3.5.2 & 3.6.1 & 3.6.4 & 3.6.5 & 4.1.1 & 4.2.1 & 4.6.1 & Avg  \\
\midrule
\model-finetune &0.79&\textbf{0.77}&\textbf{0.86}&0.88&\textbf{0.48}&0.24&0.63&0.40&\textbf{0.20}&\textbf{0.68}&\textbf{0.79}&0.37&\textbf{0.89}&0.93&0.56&\textbf{0.66}\\
\rowcolor{myblue}
\model&0.79&0.76&0.62&0.88&0.47&\textbf{0.26}&\textbf{0.73}&0.40&0.14&0.66&0.78&\textbf{0.40}&0.80&0.93&\textbf{0.57}&0.65\\
\bottomrule
\end{tabular}
\vspace{-0.6em}
\caption{ESP score ($\uparrow$) of \model and \model-finetune. Avg denotes average. Finetuning leads to improvement in ESP scores.}
\label{Tab: ESP_top5_finetune}
\end{table*}

\textbf{\model excels in designing enzymes that bind their respective substrates with high affinities.}
Table~\ref{Tab: ESP_all} and \ref{Tab: docking_all} highlight the ability of \model in achieving the highest average ESP score and substrate binding affinity across 323 families. 
Notably, \model obtains an average ESP score of $0.65$, surpassing the suggested enzyme-substrate interaction threshold of 0.6 from the ESP evaluator developer. In comparison,  PROTSEED achieves 0.59, RFDiffusion+ProteinMPNN achieves 0.53, and ESM2+EGNN achieves 0.61 of average ESP score. \model's ESP score is significantly higher than all previous baselines. This is further verified by the binding affinity scores. 
The enzymes generated by our EnzyGen exhibit stronger binding affinity to their corresponding substrates than all other methods across all 30 third-level testing EC categories. 
EnzyGen gains an average improvement of $-0.87$ in binding affinity over the previous best method (RFDiffusion+ProteinMPNN), which is significant. 
These results demonstrate that enzymes designed by our \model are able to bind their corresponding substrates with high affinities.
Using the MLM pretraining, the model demonstrates further performance gains. \model-1.5 achieves superior average enzyme-substrate binding scores across both ESP and docking metrics, although its average pLDDT is marginally lower than that of \model. These results validate the effectiveness of the pretraining strategy.

\textbf{\model is able to design well-folded enzymes.} According to the findings of \citet{guo2022alphafold2} and \citet{binder2022alphafold}, high pLDDT scores (e.g., > 80) indicate high confidence of the residue structure. \model attains an average pLDDT score of $87.45$ across 323 families. It outperforms all three baselines (PROTSEED, RFDiffusion+ProteinMPNN, ESM2+EGNN).  This demonstrates the capability of our model to design enzymes with stable folding, validating the value of incorporating structural information into the enzyme design process.

\section{Analysis: Diving Deep into \model}
\label{analysis}

\begin{figure}
\begin{minipage}[t]{0.5\linewidth}
\centering
\includegraphics[width=3.6cm]{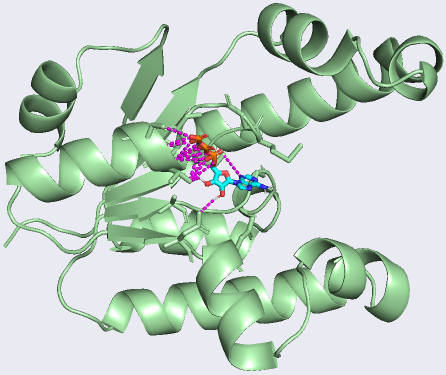}
\centerline{\small{(a) Complex of 1KAG-ATP(-4)}}
\end{minipage}%
\begin{minipage}[t]{0.5\linewidth}
\centering
\includegraphics[width=3.9cm]{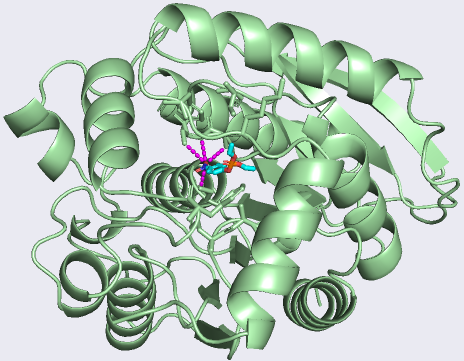}
\centerline{\small{(b) Complex of 5L2P-paraoxon}}
\end{minipage}
	\caption{Case Study: (a) Complex of designed 1KAG~(2.7.1.71, catalyzing the specific phosphorylation of the 3-hydroxyl group of shikimic acid) and substrate ATP(-4), with pLDDT=90.39, Uniprot blastp recovery rate = 58.5\%, (b) Complex of designed 5L2P~(3.1.1.2, hydrolyzing various p-nitrophenyl phosphates, aromatic esters and p-nitrophenyl fatty acids) and substrate paraoxon, with pLDDT=89.44, Uniprot blastp recovery rate = 49.4\%. Both cases show polar contacts~(hydrogen bonds) depicted in purple.} 
 \label{Fig: case_study}
\end{figure}


\begin{table*}[!t]
\scriptsize
\centering
\setlength{\tabcolsep}{1.8mm}
\begin{tabular}{lcccccccccccccccc}
\midrule
Enzyme Family & 1.1.1 & 1.11.1 & 1.14.13 & 1.14.14 & 1.2.1 & 2.1.1 & 2.3.1 & 2.4.1 & 2.4.2 & 2.5.1 & 2.6.1 & 2.7.1 & 2.7.10 & 2.7.11 & 2.7.4   \\
\midrule
\rowcolor{myblue}
\model-1.5&2.38&1.04&0.73&1.28&4.08&4.04&5.29&2.98&3.48&3.32&3.22&3.64&3.36&3.35&2.08\\
\midrule
Enzyme Family & 2.7.7 & 3.1.1 & 3.1.3 & 3.1.4 & 3.2.2 & 3.4.19 & 3.4.21 & 3.5.1 & 3.5.2 & 3.6.1 & 3.6.4 & 3.6.5 & 4.1.1 & 4.2.1 & 4.6.1 & Avg  \\
\midrule
\rowcolor{myblue}
\model-1.5&6.98&3.23&9.72&2.14&7.43&9.18&8.42&3.88&2.03&3.60&9.01&6.92&7.56&1.32&6.20& 4.40\\
\bottomrule
\end{tabular}
\vspace{-0.6em}
\caption{Self-consistency (RMSD between the folded structure of the designed sequence and the designed structure, $\downarrow$) of \model-1.5.}
\label{Tab: RMSD_consistency}
\end{table*}

\begin{figure*}[!h]
\begin{minipage}[t]{0.25\linewidth}
\centering
\includegraphics[width=3.8cm]{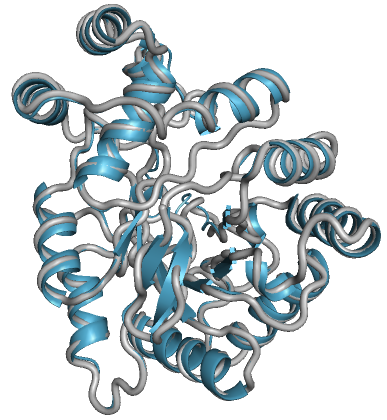}
\centerline{\small{\shortstack{(a) 1AZ2, RMSD=0.386Å, \\ EC=1.1.1.21}}}
\end{minipage}%
\begin{minipage}[t]{0.25\linewidth}
\centering
\includegraphics[width=3.8cm]{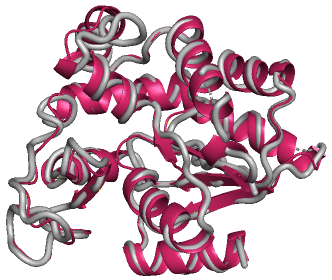}
\centerline{\small{\shortstack{(b) 1E4V, RMSD=0.619Å, \\ EC=2.7.4.3}}}
\end{minipage}%
\begin{minipage}[t]{0.25\linewidth}
\centering
\includegraphics[width=4.15cm]{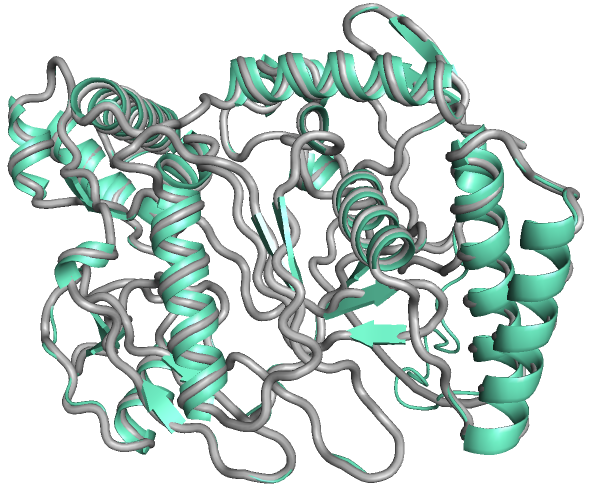}
\centerline{\small{\shortstack{ (c) 1M9Q, RMSD=0.303Å, \\EC=1.14.13.39}}}
\end{minipage}%
\begin{minipage}[t]{0.25\linewidth}
\centering
\includegraphics[width=4.05cm]{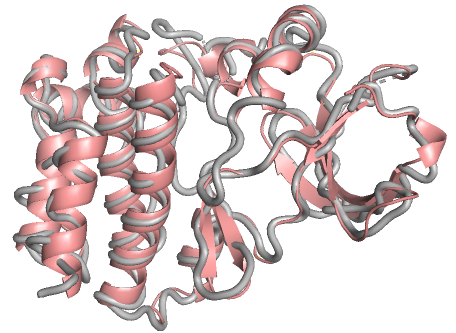}
\centerline{\small{\shortstack{(d) 4HCU, RMSD=0.560Å, \\EC=2.7.10.2}}}
\end{minipage}
	\caption{Case study of designed backbone structure aligned with ground truth.} 
 \label{Fig: case_rmsd}
\end{figure*}

\subsection{Ablation Study: Do Interleaving Network and Substrate Constraint Help?}
\textbf{They both contribute to designed enzymes with better substrate binding}.
In comparison to directly concatenating ESM2 and EGNN, Figure~\ref{Fig: ablation_interleaving} (a) illustrates that \model consistently enhances the average ESP score across 323 families when considering different candidate sets. In Figure~\ref{Fig: ablation_interleaving} (b), \model achieves higher AlphaFold2 pLDDT than ESM2+EGNN across various candidate sets. These results affirm that the interleaving network within our proposed \layer facilitates information exchange at different granularities, thereby aiding the design of well-folded enzymes with better substrate binding functions.
We further study whether the proposed enzyme-substrate interaction constraint is useful. As shown in Figure~\ref{Fig: ablation_interleaving} (c), incorporating the enzyme-substrate interaction constraint into the training process improves substrate binding affinities, boosting enzyme design with stronger substrate binding.

\subsection{Ablation Study: Does Model Scale Help?}
\textbf{\model is scalable.} To assess the scalability of our \model, we compare our \model~(714M) with a randomly initialized 12-layer model~(117M) trained on the same dataset. Figure~\ref{Fig: ablation_interleaving} (d) presents the average ESP score across 323 categories for different candidate set. The results indicate that \model attains a higher enzyme-substrate interaction score as the model scales up, and this difference becomes more obvious when more candidates are considered. This confirms the scalability advantage of our \model.

\subsection{Does Further Finetuning Bring Additional Benefits?}
\textbf{\model shows improvement with further finetuning.} Drawing inspiration from the strategy of finetuning pretrained multilingual models for specific languages~\cite{liu2020multilingual,conneau2020unsupervised}, we conduct further finetuning on each third-level category. Each category undergoes an additional finetuning of 30 epochs, leading to \model-finetune. As presented in Table~\ref{Tab: ESP_top5_finetune}, 15 categories exhibit improvements on ESP scores after finetuning. Simultaneously, the performance of 8 categories shows degradation, and it is noted that all of them have training sizes smaller than 1000, except for 2.3.1. These findings demonstrate that \model can indeed benefit from the further finetuning technique, with larger training data sizes yielding more substantial improvements.

\subsection{Zero-shot Generalization Capability}
To assess the generalization capabilities of our \model, we respectively select two unseen substrates (magnesium(2+), zinc(2+)) and two unseen fourth-level categories (2.7.13.3, 6.3.2.4) from Swiss-Prot, which are not included in our \dataset benchmark. For these enzyme cases, we utilize their structures provided by the AlphaFold Database. The ESP scores of the top-32 candidates for each new fourth-level category or substrate are presented in Figure~\ref{Fig: analysis} (a). Remarkably, the average ESP score is $0.83$ across two fourth-level categories and $0.89$ across two substrates, with more than $85\%$ of the cases scoring over 0.6. These observations confirm that functionally important sites are crucial to design functional enzymes. Even not trained on these categories, \textbf{our \model exhibits the capability to design enzymes with substrate binding ability.}

\subsection{Does \model Learn the Family Relationship?}
\textbf{Families with closer functions have closer embeddings.} We cluster the 3,157 fourth-level family category embeddings in Figure~\ref{Fig: analysis} (b). It shows that enzyme families from the same superfamily are clustered together, displaying closer tag representations in the embedding space. This observation confirms that our family tag learns useful function information which can provide guidance for the design of desirable enzymes. 


\subsection{Designing Novel Enzymes with \model}
\textbf{\model is able to design novel enzymes that bind to specific substrates.} Figure~\ref{Fig: case_study} (a) and (b) showcase two designed enzymes based on PDB 1KAG (catalyzing the specific phosphorylation of the 3-hydroxyl group of shikimic acid) and 5L2P~(hydrolyzing various p-nitrophenyl phosphates, aromatic esters and p-nitrophenyl fatty acids) respectively. 
Both of the two designed enzymes achieve pLDDT approaching or exceeding 90 and substrate binding affinities below -10. Additionally, the Gnina docked complexes show polar contacts~(hydrogen bonds) between the enzymes and substrates. This affirms the enzyme-substrate interaction function. We carry out blastp search of designed enzymes in Uniprot, yielding an amino acid identity of \textbf{58.5\%} and \textbf{49.4\%}, respectively, to the most similar enzyme. Together, these results show that our model is capable of designing novel enzymes with high substrate binding affinities. Additional cases are detailed in Appendix~\ref{Appendix: case_study}, each exhibiting a pLDDT higher than 80, a binding affinity below -10, and a Uniprot blastp identity rate lower than 65\%. Enzyme sequences for all analyzed cases are also available in Appendix Table~\ref{Tab: appendix_case_study_seq}.

\subsection{How Consistent Are the Designed Sequence and Structure?}
To assess the consistency between the designed enzyme sequences and the corresponding backbone structures, we report the RMSD between the folded structure of each designed sequence and the designed structure in Table~\ref{Tab: RMSD_consistency}. The results show that 9 out of 30 categories achieve an RMSD below 3, indicating good structural alignment. Additionally, Figure~\ref{Fig: case_rmsd} presents four representative cases, each exhibiting an RMSD below 1 between the folded and designed structures, demonstrating near-perfect alignment. These results confirm that \model-1.5 effectively designs sequences and structures that are highly consistent with one another.

\section{Discussion}
Enzymes, as genetically encoded biocatalysts, play a crucial role in accelerating chemical reactions and find extensive applications across various fields. Designing enzymes that can specifically bind to target substrates has always been a complex challenge. Our \model demonstrates a remarkable ability to design well-folded and effective enzymes that bind to specific substrates across diverse enzyme categories. Despite its impressive performance, certain limitations remain, which we will address in this section.

First, we collected enzyme data from the PDB, covering $3,157$ fourth-level enzyme classes. Although there are $8,422$ fourth-level classes in total, our dataset includes only a subset of these categories. Future work could focus on expanding this dataset by incorporating data from larger databases, such as Swiss-Prot and the AlphaFold structure database, to encompass a broader range of enzyme classes.

Second, we address the substrate-binding constraint using an enzyme-substrate binding prediction loss. While this provides a basic constraint, it is relatively weak concerning the enzyme's catalytic functions. Future research could focus on designing enzymes that bind to a specific 3D substrate structure, potentially leading to the formation of a more effective enzyme-substrate complex.

\section{Conclusion}
\label{conclusion}
This paper introduces \model, a unified generative model for designing functional enzymes across diverse families. \model simultaneously generates enzyme sequence and backbone structure guided by automatically identified functionally important sites and a given substrate. We employ a set of three losses to train \model.
To comprehensively evaluate \model, we construct \dataset, a benchmark for enzyme design, including all available enzymes within PDB across 3157 enzyme families. Experimental results demonstrate \model's capability to design well-folded enzymes with strong enzyme-substrate interaction functions. 

\section*{Acknowledgements}
This research is supported by the National Institutes of Health (R35GM147387 to Y.Y.), a seed grant from the NSF Molecule Maker Lab Institute (grant \#2019897), the UC Santa Barbara Faculty Research Grant (to L.L.), and the generous support by Pittsburgh Supercomputing Center Neocortex grant. The authors thank the anonymous reviewers and Siqi Ouyang, Yuwei Yang, Yufei Song, Jielin Qiu, and Yujia Gao for their valuable comments.

\section*{Impact Statement}
\label{impact_statement}
This paper contributes to the effective and efficient design of functional enzymes, thereby advancing the fields of functional protein design and AI for molecule design. There are many potential societal consequences of our work, none of which we feel must be specifically highlighted here.




\bibliography{icml2024}
\bibliographystyle{icml2024}

\clearpage
\appendix
\section*{Appendix}
\section{Proof: SE(3) Equivariance}
\label{proof_section_3_5}

\subsection{Proof: \layer Is SE(3) Equivariant}
\label{Appendix: theorem 3.1}
In this section we prove that our proposed \layer is translation equivariant on $\boldsymbol{x}$ for any translation vector $\boldsymbol{t}\in \mathbb{R}^3$ and rotation equivariant for any $R$ from SO(3) group. More formally, we will prove the \layer satisfies:
\begin{equation}
\small 
\boldsymbol{H}^{l+1}, R\boldsymbol{x}^{l+1}+\boldsymbol{t} = \mathrm{\layer}(\boldsymbol{H}^{l}, R\boldsymbol{x}^{l}+\boldsymbol{t})
\end{equation}

When $l=0$:\\
For message update, the distance between two residues is invariant as $d_{ij}^{l} = ||R\boldsymbol{x}_i^{l}+\boldsymbol{t}-(R\boldsymbol{x}^{l}_j+\boldsymbol{t})||^2=(\boldsymbol{x}_i^{l}-\boldsymbol{x}^{l}_j)^TR^TR(\boldsymbol{x}_i^{l}-\boldsymbol{x}^{l}_j)=||\boldsymbol{x}_i^{l}-\boldsymbol{x}^{l}_j||^2$. 
$\boldsymbol{h}^{l}_i$ is the embedding of residue or [mask] token, and thus it is always invariant.
Therefore, $\boldsymbol{h}^{l+0.5}_i$ is also invariant.
Since $\boldsymbol{h}^{l+0.5}_i$, $\boldsymbol{h}^{l+0.5}_j$ and $d_{ij}^{l}$ are all invariant, and thus $\boldsymbol{m}_{ij}^{l+1}$ is also invariant to translation $\boldsymbol{t}$ and rotation $R$ on $\boldsymbol{x}$.

For coordinate update, updated $\boldsymbol{x}^{l+1}$ is equivariant to the translation $\boldsymbol{t}$ and rotation $R$ on input $\boldsymbol{x}^l$:
\begin{equation}
\small
\begin{split}
&(R\boldsymbol{x}^{l}_i+\boldsymbol{t})+\sum_{j\in \mathrm{Neighbor(i)}}(R\boldsymbol{x}^{l}_i+\boldsymbol{t}-(R\boldsymbol{x}^{l}_j+\boldsymbol{t}))\cdot \mathrm{FFN}(\boldsymbol{m}_{ij}^{l+1})\\
&=R(\boldsymbol{x}^{l}_i+\sum_{j\in \mathrm{Neighbor(i)}}(\boldsymbol{x}^{l}_i-\boldsymbol{x}^{l}_j)\cdot \mathrm{FFN}(\boldsymbol{m}_{ij}^{l+1})) + \boldsymbol{t} \\
&=R\boldsymbol{x}^{l+1}_i + \boldsymbol{t}
\end{split}
\end{equation}

For residue update, $\boldsymbol{m}_{ij}^{l+1}$ and $\boldsymbol{h}_i^{l+0.5}$ is invariant to translation $\boldsymbol{t}$ and rotation $R$ on $\boldsymbol{x}$, so $\boldsymbol{h}_i^{l+1}$ is also invariant to translation $\boldsymbol{t}$ and rotation $R$ on $\boldsymbol{x}$.

When $l>=1$:\\
We have proved that when $l=0$, $\boldsymbol{h}^1_i$ is invariant to rotation and translation on $\boldsymbol{x}$. Taking $\boldsymbol{H}^1$ as the second layer input and following the above process, we can prove $\boldsymbol{h}^2_i$ is also invariant. Repeating this process from $l=1$ to $L-1$, we can get the same conclusion.

Combining the above two scenarios together, we have $\boldsymbol{H}^{l+1}, R\boldsymbol{x}^{l+1}+\boldsymbol{t} = \mathrm{\layer}(\boldsymbol{H}^{l}, R\boldsymbol{x}^{l}+\boldsymbol{t})$ for $l=0$ to $L-1$. Therefore, our proposed \layer is $SE(3)$-equivariant.

\begin{table*}[ht]
\footnotesize
\begin{center}
\begin{tabular}{lc}
\midrule
Third-Level Category  & Function \\
\midrule
1.1.1 &  With NAD+ or NADP+ as acceptor \\
1.11.1 & Peroxidases\\
1.14.13 & With NADH or NADPH as one donor, and incorporation  \\
&of one atom of oxygen into the other donor \\
1.14.14 & With reduced flavin or flavoprotein as one donor\\
& and incorporation of one atom of oxygen into the other donor \\
1.2.1 & With NAD+ or NADP+ as acceptor \\
2.1.1 & Methyltransferases\\
2.3.1 & Transferring groups other than aminoacyl groups\\
2.4.1 & Hexosyltransferases \\
2.4.2 & Pentosyltransferases \\
2.5.1 & Transferring alkyl or aryl groups, other than methyl groups (only sub-subclass identified to date) \\
2.6.1 & Transaminases \\
2.7.1 & Phosphotransferases with an alcohol group as acceptor \\
2.7.10 & Protein-tyrosine kinases \\
2.7.11 & Protein-serine/threonine kinases\\
2.7.4 & Phosphotransferases with a phosphate group as acceptor \\
2.7.7 & Nucleotidyltransferases \\
3.1.1 & Carboxylic-ester hydrolases \\
3.1.3 & Phosphoric-monoester hydrolases \\
3.1.4 & Phosphoric-diester hydrolases \\
3.2.2 & Hydrolysing N-glycosyl compounds \\
3.4.19 & Omega peptidases \\
3.4.21 & Serine endopeptidases \\
3.5.1 & In linear amides \\
3.5.2 &  In cyclic amides \\
3.6.1 &  In phosphorus-containing anhydrides \\
3.6.4 & Acting on acid anhydrides to facilitate cellular and subcellular movement \\
3.6.5 & Acting on GTP to facilitate cellular and subcellular movement \\
4.1.1 &  Carboxy-lyases \\
4.2.1 & Hydro-lyases \\
4.6.1 & Phosphorus-oxygen lyases (only sub-subclass identified to date) \\
\bottomrule
\end{tabular}
\end{center}
\caption{Third-level category information for validation and test sets.}
\label{Tab: test_category}
\end{table*}

\subsection{Proof: \model Is SE(3) Equivariant}
\label{proof: corollary_3_2}

We provide the proof of corollary 3.2 as follows:
\begin{proof}\renewcommand{\qedsymbol}{}
\begin{equation}
\small
\begin{split}
&\mathrm{\model}(\boldsymbol{H}^{0}, R\boldsymbol{x}^{0}+\boldsymbol{t}) \\&= \mathrm{\layer}^{L-1}\circ \mathrm{\layer}^{L-2}\circ \cdot \cdot \cdot \circ \mathrm{\layer}^{0}(\boldsymbol{H}^{0}, R\boldsymbol{x}^{0}+\boldsymbol{t}) ) \\
&= \mathrm{\layer}^{L-1}\circ \mathrm{\layer}^{L-2}\circ \cdot \cdot \cdot \circ \mathrm{\layer}^{1}(\boldsymbol{H}^{1}, R\boldsymbol{x}^{1}+\boldsymbol{t})) \\
&=...\\
&=\mathrm{\layer}^{L-1}(\boldsymbol{H}^{L-1}, R\boldsymbol{x}^{L-1}+\boldsymbol{t})) = \boldsymbol{H}^{L}, R\boldsymbol{x}^{L}+\boldsymbol{t}
\end{split}
\end{equation} 
\end{proof}



\begin{table*}[!t]
\tiny
\centering
\setlength{\tabcolsep}{0.6mm}
\begin{tabular}{lcccccccccccccccc}
\midrule
Enzyme Family & 1.1.1.239&1.1.1.270&1.1.1.201&1.1.1.184&1.1.1.25&1.1.1.62&1.1.1.197&1.1.1.35&1.1.1.248&1.1.1.271&1.1.1.372&1.1.1.357&1.1.1.27&1.1.1.286&1.1.1.47&1.1.1.3  \\
\rowcolor{myblue}
\model & 0.62&0.15&0.09&0.16&0.92&0.38&0.39&0.55&0.89&0.62&0.86&0.71&0.98&0.14&0.99&0.39 \\
\midrule
Enzyme Family & 1.1.1.56&1.1.1.105&1.1.1.77&1.1.1.34&1.1.1.399&1.1.1.42&1.1.1.49&1.1.1.44&1.1.1.40&1.11.1.24&1.11.1.12&1.11.1.15&1.11.1.9&1.11.1.11&1.11.1.5&1.11.1.6  \\
\rowcolor{myblue}
\model & 0.39&0.64&0.92&0.07&0.98&0.65&0.82&0.16&0.99&1.0&1.0&1.0&0.99&0.92&0.99&0.9\\
\midrule
Enzyme Family & 1.11.1.21&1.14.13.39&1.14.13.9&1.14.13.23&1.14.14.18&1.14.14.47&1.14.14.25&1.14.14.154&1.14.14.1&1.14.14.32&1.14.14.24&1.14.14.26&1.14.14.91&1.2.1.3&1.2.1.50&1.2.1.44 \\
\rowcolor{myblue}
\model & 0.99&0.99&0.4&0.19&0.6&0.27&0.24&0.03&0.56&0.36&0.99&0.22&0.04&0.44&0.66&0.48\\
\midrule
Enzyme Family & 1.2.1.12&1.2.1.59&1.2.1.13&1.2.1.11&1.2.1.10&1.2.1.5&1.2.1.47&1.2.1.36&1.2.1.9&1.2.1.26&1.2.1.88&1.2.1.84&1.2.1.89&1.2.1.90&2.1.1.309&2.1.1.37 \\
\rowcolor{myblue}
\model &0.94&0.84&0.36&0.92&0.59&0.96&0.87&0.6&0.84&0.99&0.73&0.03&0.82&0.91&0.06&0.94 \\
\midrule
Enzyme Family & 2.1.1.364&2.1.1.221&2.1.1.348&2.1.1.320&2.1.1.6&2.1.1.366&2.1.1.367&2.1.1.244&2.1.1.359&2.1.1.72&2.1.1.220&2.1.1.45&2.1.1.98&2.1.1.79&2.1.1.8&2.1.1.233 \\
\rowcolor{myblue}
\model &0.73&0.02&1.0&0.76&1.0&0.84&0.99&1.0&0.95&0.82&1.0&0.85&0.78&0.97&0.02&0.02\\
\midrule
Enzyme Family & 2.1.1.319&2.1.1.5&2.1.1.354&2.1.1.173&2.3.1.48&2.3.1.57&2.3.1.82&2.3.1.258&2.3.1.87&2.3.1.108&2.3.1.39&2.3.1.286&2.3.1.94&2.3.1.117&2.3.1.180&2.3.1.16 \\
\rowcolor{myblue}
\model &0.49&0.72&0.82&0.76&0.75&0.34&0.97&0.35&0.07&0.2&0.43&0.31&0.67&0.53&0.98&0.79 \\
\midrule
Enzyme Family & 2.3.1.97&2.3.1.9&2.3.1.47&2.3.1.26&2.3.1.50&2.3.1.37&2.4.1.186&2.4.1.135&2.4.1.38&2.4.1.255&2.4.1.143&2.4.1.264&2.4.1.182&2.4.1.345&2.4.1.115&2.4.1.15 \\
\rowcolor{myblue}
\model &0.76&0.77&0.32&0.97&0.35&0.07&0.35&0.68&0.14&0.19&0.11&0.43&0.48&0.52&0.32&0.35\\
\midrule
Enzyme Family & 2.4.1.21&2.4.1.41&2.4.1.155&2.4.1.11&2.4.1.225&2.4.1.1&2.4.2.30&2.4.2.2&2.4.2.8&2.4.2.3&2.4.2.1&2.4.2.19&2.4.2.17&2.4.2.21&2.4.2.64&2.4.2.14 \\
\rowcolor{myblue}
\model &0.98&0.96&0.29&0.38&0.56&0.84&0.89&0.49&0.93&0.99&0.98&0.93&0.92&0.69&0.65&0.15\\
\midrule
Enzyme Family & 2.4.2.36&2.5.1.18&2.5.1.31&2.5.1.68&2.5.1.55&2.5.1.16&2.5.1.1&2.5.1.15&2.5.1.79&2.5.1.87&2.5.1.29&2.5.1.59&2.5.1.126&2.5.1.47&2.5.1.46&2.5.1.54 \\
\rowcolor{myblue}
\model &0.68&0.27&0.28&0.99&0.64&0.91&0.99&0.58&0.88&0.12&0.99&0.98&0.82&1.0&0.96&0.95\\
\midrule
Enzyme Family & 2.5.1.6&2.5.1.7&2.5.1.19&2.5.1.26&2.6.1.21&2.6.1.9&2.6.1.16&2.6.1.99&2.6.1.51&2.6.1.63&2.6.1.3&2.6.1.1&2.6.1.57&2.6.1.7&2.6.1.81&2.6.1.73 \\
\rowcolor{myblue}
\model &0.98&0.98&0.98&0.95&0.05&0.23&0.32&0.06&0.7&0.99&0.99&0.41&0.31&1.0&0.29&0.94\\
\midrule
Enzyme Family & 2.6.1.83&2.6.1.62&2.6.1.85&2.6.1.82&2.7.1.107&2.7.1.154&2.7.1.26&2.7.1.71&2.7.1.21&2.7.1.24&2.7.1.113&2.7.1.153&2.7.1.23&2.7.1.35&2.7.1.33&2.7.1.68\\
\rowcolor{myblue}
\model &0.95&0.71&0.04&0.66&0.39&0.83&0.06&0.99&0.75&0.14&0.57&0.61&0.94&0.97&0.81&0.98\\
\midrule
Enzyme Family & 2.7.1.20&2.7.1.178&2.7.1.19&2.7.1.159&2.7.1.82&2.7.1.73&2.7.1.40&2.7.1.1&2.7.1.47&2.7.1.158&2.7.1.5&2.7.10.2&2.7.10.1&2.7.11.1&2.7.11.21&2.7.11.25 \\
\rowcolor{myblue}
\model &0.68&0.57&0.34&0.09&0.78&0.13&0.99&0.98&0.89&0.99&0.98&0.88&0.98&0.92&0.63&1.0\\
\midrule
Enzyme Family & 2.7.11.22&2.7.11.30&2.7.11.24&2.7.11.26&2.7.11.2&2.7.11.16&2.7.4.6&2.7.4.3&2.7.4.8&2.7.4.14&2.7.4.9&2.7.4.25&2.7.4.10&2.7.4.22&2.7.4.24&2.7.4.1 \\
\rowcolor{myblue}
\model &1.0&1.0&1.0&1.0&0.99&1.0&1.0&0.75&0.4&0.99&0.96&0.94&0.7&0.24&0.15&0.39\\
\midrule
Enzyme Family & 2.7.7.7&2.7.7.6&2.7.7.65&2.7.7.60&2.7.7.86&2.7.7.31&2.7.7.12&2.7.7.52&2.7.7.27&2.7.7.23&2.7.7.9&2.7.7.48&3.1.1.4&3.1.1.3&3.1.1.78&3.1.1.2 \\
\rowcolor{myblue}
\model &0.91&0.29&0.42&0.8&0.38&0.98&0.98&0.41&0.88&0.15&0.57&0.38&0.99&0.19&0.51&0.84\\
\midrule
Enzyme Family & 3.1.1.23&3.1.1.72&3.1.1.47&3.1.1.34&3.1.1.8&3.1.1.7&3.1.1.5&3.1.3.32&3.1.3.46&3.1.3.3&3.1.3.104&3.1.3.16&3.1.3.108&3.1.3.7&3.1.3.11&3.1.3.36 \\
\rowcolor{myblue}
\model &0.5&0.43&0.96&0.86&0.86&0.91&0.13&0.17&0.67&0.5&0.07&0.83&0.99&0.23&0.93&0.53\\
\midrule
Enzyme Family & 3.1.3.10&3.1.3.76&3.1.3.1&3.1.4.52&3.1.4.17&3.1.4.55&3.1.4.35&3.1.4.53&3.2.2.6&3.2.2.9&3.2.2.16&3.2.2.8&3.2.2.3&3.2.2.1&3.2.2.10&3.4.19.12 \\
\rowcolor{myblue}
\model &0.25&0.21&0.09&0.23&0.95&0.03&0.98&0.99&0.41&0.82&0.2&0.55&0.51&0.87&0.77&0.44\\
\midrule
Enzyme Family & 3.4.19.5&3.4.19.14&3.4.19.13&3.4.21.53&3.4.21.92&3.4.21.107&3.4.21.22&3.5.1.98&3.5.1.26&3.5.1.97&3.5.1.124&3.5.1.9&3.5.1.44&3.5.1.1&3.5.1.2&3.5.1.25 \\
\rowcolor{myblue}
\model &0.22&0.07&0.07&0.73&0.69&0.91&0.57&0.23&0.08&0.99&0.83&0.32&0.56&0.07&0.07&0.41\\
\midrule
Enzyme Family & 3.5.1.16&3.5.1.92&3.5.1.99&3.5.1.62&3.5.1.23&3.5.2.17&3.5.2.6&3.5.2.10&3.6.1.12&3.6.1.23&3.6.1.55&3.6.1.56&3.6.1.9&3.6.1.60&3.6.1.1&3.6.1.72 \\
\rowcolor{myblue}
\model &0.13&0.11&0.22&0.44&0.27&0.15&0.13&0.63&0.97&0.11&0.63&0.45&0.88&0.93&0.9&0.54\\
\midrule
Enzyme Family & 3.6.1.15&3.6.1.58&3.6.1.8&3.6.1.22&3.6.1.27&3.6.1.29&3.6.1.5&3.6.4.6&3.6.4.12&3.6.4.13&3.6.4.10&3.6.5.2&3.6.5.1&3.6.5.5&4.1.1.23&4.1.1.85 \\
\rowcolor{myblue}
\model &0.63&0.97&0.99&0.84&0.75&0.14&0.76&0.73&0.79&0.4&0.96&0.4&0.42&0.29&0.97&0.87\\
\midrule
Enzyme Family & 4.1.1.17&4.1.1.37&4.1.1.39&4.1.1.28&4.1.1.12&4.1.1.49&4.1.1.25&4.1.1.32&4.2.1.10&4.2.1.19&4.2.1.1&4.2.1.24&4.2.1.22&4.2.1.8&4.2.1.2&4.6.1.1 \\
\rowcolor{myblue}
\model &0.02&0.57&0.88&0.02&0.33&0.99&0.04&0.45&0.81&0.85&1.0&0.64&0.01&0.81&0.96&0.65\\
\midrule
Enzyme Family & 4.6.1.17&4.6.1.12&4.6.1.2&&&&&&&&&&&&&Avg \\
\rowcolor{myblue}
\model &0.49&0.4&0.52&&&&&&&&&&&&&0.65\\
\bottomrule
\end{tabular}
\vspace{-0.6em}
\caption{Enzyme-substrate ESP score~($\uparrow$) for the 323 testing fourth-level categories. Note 0.6 or higher ESP score indicates a positive binding. Avg denotes average.}
\label{Tab: esp_299families}
\end{table*}

\begin{figure*}
  \centering
  \includegraphics[width=16.0cm]{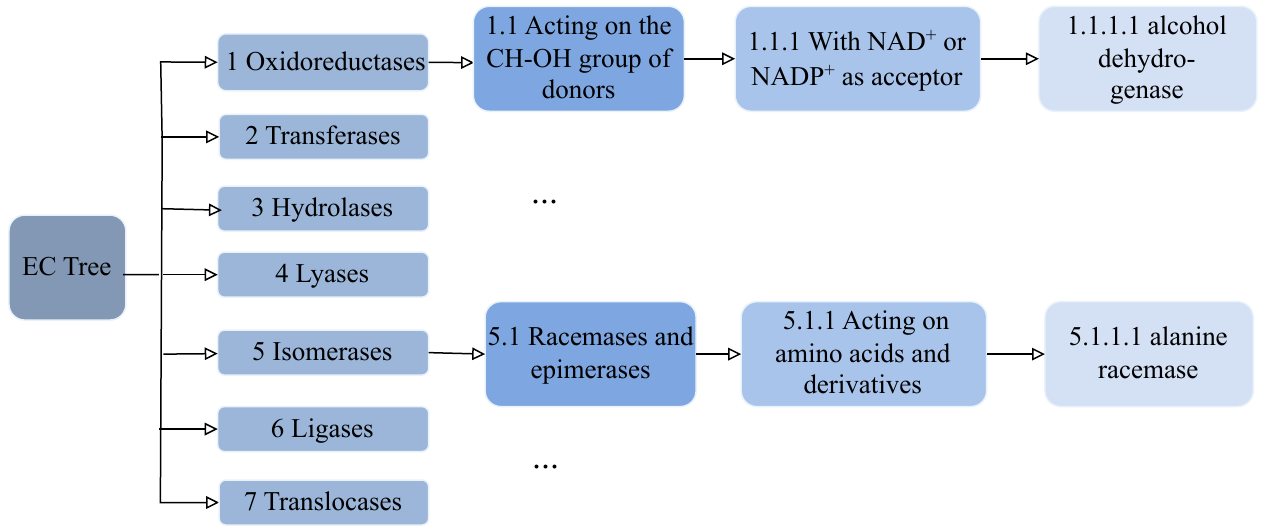}
  \caption{Enzyme Classification~(EC) Tree in BRENDA.}
  \label{Fig: appendix_ec_tree}
\end{figure*}

\begin{table*}[!t]
\scriptsize
\centering
\setlength{\tabcolsep}{1.8mm}
\begin{tabular}{lcccccccccccccccc}
\midrule
Enzyme Family & 1.1.1 & 1.11.1 & 1.14.13 & 1.14.14 & 1.2.1 & 2.1.1 & 2.3.1 & 2.4.1 & 2.4.2 & 2.5.1 & 2.6.1 & 2.7.1 & 2.7.10 & 2.7.11 & 2.7.4 & Avg  \\
\midrule
PROTSEED &  0.98&0.99&0.48&0.98&0.98&0.99&0.99&0.99&0.99&0.99&0.99&0.99&0.99&0.99&0.99 &  -- \\
RFDiffusion+IF & 0.89&0.95&0.72&0.58&0.95&0.68&0.76&0.95&0.86&0.95&0.95&0.95&0.82&0.95&0.95 & -- \\
ESM2+EGNN &0.96&0.97&0.69&0.97&0.98&0.98&0.98&0.97&0.97&0.97&0.97&0.97&0.98&0.98&0.98 & -- \\
\rowcolor{myblue}
\model & 0.99&1.00&0.73&1.00&1.00&1.00&1.00&0.98&1.00&1.00&1.00&1.00&1.00&1.00&1.00 \\
\midrule
Enzyme Family & 2.7.7 & 3.1.1 & 3.1.3 & 3.1.4 & 3.2.2 & 3.4.19 & 3.4.21 &  3.5.1 & 3.5.2 & 3.6.1 & 3.6.4 & 3.6.5 & 4.1.1 & 4.2.1 & 4.6.1 & Avg  \\
\midrule
PROTSEED & 0.98&0.93&0.95&0.98&0.92&0.91&0.99&0.99&0.25&0.97&0.98&0.47&0.95&0.99&0.97&0.92  \\
RFDiffusion+IF & 0.98&0.97&0.93&0.94&0.92&0.94&0.97&0.96&0.60&0.97&0.97&0.94&0.92&0.98&0.97&0.94 \\
ESM2+EGNN &0.98&0.97&0.93&0.94&0.92&0.94&0.97&0.96&0.60&0.97&0.97&0.94&0.92&0.98&0.97&0.94 \\
\rowcolor{myblue}
\model & 1.00&1.00&1.00&1.00&0.96&0.99&1.00&0.99&0.63&0.99&1.00&0.89&0.99&1.00&0.99&\textbf{0.97}\\
\bottomrule
\end{tabular}
\vspace{-0.6em}
\caption{Enzyme-substrate ESP score~($\uparrow$) of the top-1 candidate. IF denotes the inverse folding model ProteinMPNN. Note 0.6 or higher ESP score indicates a positive binding. Avg denotes average.}
\label{Tab: ESP_top1}
\end{table*}

\section{Enzyme Classification Tree}
\label{ec_tree_appendix}

We provide an illustration of enzyme classification~(EC) tree in figure~\ref{Fig: appendix_ec_tree}. These categories are applied by our \model to guide the enzyme design with specific functions.

\section{Data Statistics}

\label{appendix:data_statistics}
We provide the third-level category information in the validation and test sets in Table~\ref{Tab: test_category}.


\begin{table*}[!t]
\tiny
\centering
\setlength{\tabcolsep}{0.6mm}
\begin{tabular}{lcccccccccccccccc}
\midrule
Enzyme Family & 1.1.1.239&1.1.1.270&1.1.1.201&1.1.1.184&1.1.1.25&1.1.1.62&1.1.1.197&1.1.1.35&1.1.1.248&1.1.1.271&1.1.1.372&1.1.1.357&1.1.1.27&1.1.1.286&1.1.1.47&1.1.1.3 \\
\rowcolor{myblue}
\model & -4.29&-9.76&-11.74&-12.03&-3.29&-13.14&-3.32&-12.46&-4.75&-3.52&-3.29&-10.41&-8.47&-6.32&-12.56&-6.57 \\
\midrule
Enzyme Family & 1.1.1.56&1.1.1.105&1.1.1.77&1.1.1.34&1.1.1.399&1.1.1.42&1.1.1.49&1.1.1.44&1.1.1.40&1.11.1.24&1.11.1.12&1.11.1.15&1.11.1.9&1.11.1.11&1.11.1.5&1.11.1.6  \\
\rowcolor{myblue}
\model & -12.31&-13.01&-10.87&-9.23&-10.65&-9.37&-9.51&-9.01&-9.87&-5.03&-4.42&-4.48&-4.42&-4.38&-4.55&-5.44\\
\midrule
Enzyme Family & 1.11.1.21&1.14.13.39&1.14.13.9&1.14.13.23&1.14.14.18&1.14.14.47&1.14.14.25&1.14.14.154&1.14.14.1&1.14.14.32&1.14.14.24&1.14.14.26&1.14.14.91&1.2.1.3&1.2.1.50&1.2.1.44  \\
\rowcolor{myblue}
\model & -6.64&-5.1&-6.24&-6.92&-12.34&-10.96&-8.63&-10.25&-7.32&-10.12&-8.76&-9.04&-9.84&-5.64&-5.66&-5.65\\
\midrule
Enzyme Family & 1.2.1.12&1.2.1.59&1.2.1.13&1.2.1.11&1.2.1.10&1.2.1.5&1.2.1.47&1.2.1.36&1.2.1.9&1.2.1.26&1.2.1.88&1.2.1.84&1.2.1.89&1.2.1.90&2.1.1.309&2.1.1.37 \\
\rowcolor{myblue}
\model &-6.17&-14.72&-6.22&-10.11&-10.98&-7.9&-9.44&-6.31&-12.57&-13.56&-11.75&-7.13&-8.62&-10.26&-9.11&-9.84 \\
\midrule
Enzyme Family & 2.1.1.364&2.1.1.221&2.1.1.348&2.1.1.320&2.1.1.6&2.1.1.366&2.1.1.367&2.1.1.244&2.1.1.359&2.1.1.72&2.1.1.220&2.1.1.45&2.1.1.98&2.1.1.79&2.1.1.8&2.1.1.233 \\
\rowcolor{myblue}
\model &-9.04&-8.16&-10.96&-9.51&-11.43&-9.98&-8.9&-9.01&-9.87&-9.24&-10.81&-9.64&-8.07&-7.94&-10.94&-11.95\\
\midrule
Enzyme Family & 2.1.1.319&2.1.1.5&2.1.1.354&2.1.1.173&2.3.1.48&2.3.1.57&2.3.1.82&2.3.1.258&2.3.1.87&2.3.1.108&2.3.1.39&2.3.1.286&2.3.1.94&2.3.1.117&2.3.1.180&2.3.1.16 \\
\rowcolor{myblue}
\model &-5.93&-7.97&-8.63&-8.9&-10.63&-6.52&-8.29&-9.47&-9.27&-9.41&-9.61&-8.76&-10.12&-10.59&-10.91&-9.98\\
\midrule
Enzyme Family & 2.3.1.97&2.3.1.9&2.3.1.47&2.3.1.26&2.3.1.50&2.3.1.37&2.4.1.186&2.4.1.135&2.4.1.38&2.4.1.255&2.4.1.143&2.4.1.264&2.4.1.182&2.4.1.345&2.4.1.115&2.4.1.15 \\
\rowcolor{myblue}
\model &-7.56&-7.21&-8.34&-10.25&-9.61&-8.9&-8.61&-8.67&-10.83&-11.73&-5.25&-5.93&-6.35&-8.76&-9.44&-10.25 \\
\midrule
Enzyme Family & 2.4.1.21&2.4.1.41&2.4.1.155&2.4.1.11&2.4.1.225&2.4.1.1&2.4.2.30&2.4.2.2&2.4.2.8&2.4.2.3&2.4.2.1&2.4.2.19&2.4.2.17&2.4.2.21&2.4.2.64&2.4.2.14 \\
\rowcolor{myblue}
\model &-11.67&-10.1&-11.32&-11.53&-10.54&-11.84&-11.13&-13.41&-13.88&-9.98&-8.24&-10.61&-9.01&-9.44&-15.76&-13.06 \\
\midrule
Enzyme Family & 2.4.2.36&2.5.1.18&2.5.1.31&2.5.1.68&2.5.1.55&2.5.1.16&2.5.1.1&2.5.1.15&2.5.1.79&2.5.1.87&2.5.1.29&2.5.1.59&2.5.1.126&2.5.1.47&2.5.1.46&2.5.1.54 \\
\rowcolor{myblue}
\model &-10.64&-10.04&-11.41&-8.84&-8.79&-8.27&-8.21&-9.03&-9.98&-10.67&-12.95&-9.85&-9.87&-10.86&-11.23&-10.01\\
\midrule
Enzyme Family & 2.5.1.6&2.5.1.7&2.5.1.19&2.5.1.26&2.6.1.21&2.6.1.9&2.6.1.16&2.6.1.99&2.6.1.51&2.6.1.63&2.6.1.3&2.6.1.1&2.6.1.57&2.6.1.7&2.6.1.81&2.6.1.73 \\
\rowcolor{myblue}
\model &-5.93&-5.82&-8.63&-9.98&-7.08&-8.8&-8.98&-7.43&-7.02&-8.01&-7.81&-8.36&-8.9&-8.02&-7.65&-8.9\\
\midrule
Enzyme Family & 2.6.1.83&2.6.1.62&2.6.1.85&2.6.1.82&2.7.1.107&2.7.1.154&2.7.1.26&2.7.1.71&2.7.1.21&2.7.1.24&2.7.1.113&2.7.1.153&2.7.1.23&2.7.1.35&2.7.1.33&2.7.1.68 \\
\rowcolor{myblue}
\model &-8.01&-8.87&-8.12&-9.23&-9.21&-10.64&-12.58&-9.09&-12.07&-12.07&-9.76&-9.23&-9.65&-10.5&-10.76&-10.87\\
\midrule
Enzyme Family & 2.7.1.20&2.7.1.178&2.7.1.19&2.7.1.159&2.7.1.82&2.7.1.73&2.7.1.40&2.7.1.1&2.7.1.47&2.7.1.158&2.7.1.5&2.7.10.2&2.7.10.1&2.7.11.1&2.7.11.21&2.7.11.25 \\
\rowcolor{myblue}
\model &-8.28&-10.4&-12.43&-10.86&-11.56&-8.01&-10.87&-10.04&-10.85&-11.45&-11.86&-11.65&-9.44&-12.32&-13.13&-11.44\\
\midrule
Enzyme Family & 2.7.11.22&2.7.11.30&2.7.11.24&2.7.11.26&2.7.11.2&2.7.11.16&2.7.4.6&2.7.4.3&2.7.4.8&2.7.4.14&2.7.4.9&2.7.4.25&2.7.4.10&2.7.4.22&2.7.4.24&2.7.4.1 \\
\rowcolor{myblue}
\model &-10.32&-12.53&-13.65&-11.02&-10.83&-12.36&-11.24&-8.76&-9.89&-13.56&-13.36&-10.32&-11.96&-8.99&-8.65&-13.56\\
\midrule
Enzyme Family & 2.7.7.7&2.7.7.6&2.7.7.65&2.7.7.60&2.7.7.86&2.7.7.31&2.7.7.12&2.7.7.52&2.7.7.27&2.7.7.23&2.7.7.9&2.7.7.48&3.1.1.4&3.1.1.3&3.1.1.78&3.1.1.2 \\
\rowcolor{myblue}
\model &-9.02&-8.49&-8.8&-8.35&-8.02&-7.32&-7.02&-7.94&-6.53&-8.48&-7.03&-3.12&-7.79&-8.38&-5.88&-3.32\\
\midrule
Enzyme Family & 3.1.1.23&3.1.1.72&3.1.1.47&3.1.1.34&3.1.1.8&3.1.1.7&3.1.1.5&3.1.3.32&3.1.3.46&3.1.3.3&3.1.3.104&3.1.3.16&3.1.3.108&3.1.3.7&3.1.3.11&3.1.3.36 \\
\rowcolor{myblue}
\model &-5.84&-5.58&-5.44&-5.23&-7.86&-7.33&-12.05&-10.07&-8.25&-6.45&-9.1&-9.36&-9.25&-9.04&-9.13&-9.07\\
\midrule
Enzyme Family & 3.1.3.10&3.1.3.76&3.1.3.1&3.1.4.52&3.1.4.17&3.1.4.55&3.1.4.35&3.1.4.53&3.2.2.6&3.2.2.9&3.2.2.16&3.2.2.8&3.2.2.3&3.2.2.1&3.2.2.10&3.4.19.12 \\
\rowcolor{myblue}
\model &-8.12&-8.42&-7.95&-12.09&-11.35&-11.39&-12.14&-11.86&-10.5&-10.46&-10.56&-9.63&-5.2&-10.93&-10.76&-10.83\\
\midrule
Enzyme Family & 3.4.19.5&3.4.19.14&3.4.19.13&3.4.21.53&3.4.21.92&3.4.21.107&3.4.21.22&3.5.1.98&3.5.1.26&3.5.1.97&3.5.1.124&3.5.1.9&3.5.1.44&3.5.1.1&3.5.1.2&3.5.1.25 \\
\rowcolor{myblue}
\model &-10.2&-9.25&-11.56&-11.95&-10.34&-8.32&-9.87&-5.24&-3.97&-4.24&-7.37&-4.79&-8.98&-6.58&-6.64&-9.46\\
\midrule
Enzyme Family & 3.5.1.16&3.5.1.92&3.5.1.99&3.5.1.62&3.5.1.23&3.5.2.17&3.5.2.6&3.5.2.10&3.6.1.12&3.6.1.23&3.6.1.55&3.6.1.56&3.6.1.9&3.6.1.60&3.6.1.1&3.6.1.72 \\
\rowcolor{myblue}
\model &-9.01&-6.92&-4.23&-6.52&-5.24&-5.7&-6.63&-12.82&-8.7&-10.37&-11.7&-7.86&-4.62&-8.61&-9.44&-8.32\\
\midrule
Enzyme Family & 3.6.1.15&3.6.1.58&3.6.1.8&3.6.1.22&3.6.1.27&3.6.1.29&3.6.1.5&3.6.4.6&3.6.4.12&3.6.4.13&3.6.4.10&3.6.5.2&3.6.5.1&3.6.5.5&4.1.1.23&4.1.1.85 \\
\rowcolor{myblue}
\model &-9.01&-9.87&-9.23&-9.65&-8.52&-8.25&-8.53&-11.12&-13.85&-13.82&-13.41&-11.35&-11.89&-12.56&-11.04&-10.94\\
\midrule
Enzyme Family & 4.1.1.17&4.1.1.37&4.1.1.39&4.1.1.28&4.1.1.12&4.1.1.49&4.1.1.25&4.1.1.32&4.2.1.10&4.2.1.19&4.2.1.1&4.2.1.24&4.2.1.22&4.2.1.8&4.2.1.2&4.6.1.1 \\
\rowcolor{myblue}
\model &-10.17&-10.92&-11.53&-10.36&-10.63&-10.35&-6.32&-7.35&-5.41&-5.44&-5.53&-6.99&-8.65&-9.34&-4.23&-10.7\\
\midrule
Enzyme Family & 4.6.1.17&4.6.1.12&4.6.1.2&&&&&&&&&&&&&Avg \\
\rowcolor{myblue}
\model &-10.02&-11.76&-10.23&&&&&&&&&&&&&-9.44\\
\bottomrule
\end{tabular}
\vspace{-0.6em}
\caption{Substrate binding affinity~(kcal/mol, $\downarrow$) for the 323 testing fourth-level categories. Avg denotes average.}
\label{Tab: docking_299families}
\end{table*}

\begin{table*}[!t]
\scriptsize
\centering
\setlength{\tabcolsep}{1.60mm}
\begin{tabular}{lcccccccccccccccc}
\midrule
Enzyme Family & 1.1.1 & 1.11.1 & 1.14.13 & 1.14.14 & 1.2.1 & 2.1.1 & 2.3.1 & 2.4.1 & 2.4.2 & 2.5.1 & 2.6.1 & 2.7.1 & 2.7.10 & 2.7.11 & 2.7.4 & Avg  \\
\midrule
PROTSEED &  -8.37&-6.90&-7.08&-11.69&-6.61&-7.94&-12.26&-11.83&-12.93&-10.09&-8.30&-8.15&-10.09&-12.70&-10.01&-- \\
RFDiffusion+IF & -6.54&-7.39&-7.09&-11.63&-11.89&-12.59&-12.39&-9.49&-14.11&-7.87&-8.81&-10.99&-14.19&-14.71&-11.53 & --   \\
ESM2+EGNN &-12.93&-10.03&-8.74&-10.39&-8.20&-14.38&-12.40&-11.41&-14.42&-12.00&-11.01&-8.74&-13.46&-13.32&-12.23 & -- \\
\rowcolor{myblue}
\model & -16.63&-7.20&-6.85&-13.03&-14.72&-12.43&-14.21&-11.84&-16.31&-15.36&-9.23&-13.31&-12.92&-15.01&-13.56\\
\midrule
Enzyme Family & 2.7.7 & 3.1.1 & 3.1.3 & 3.1.4 & 3.2.2 & 3.4.19 & 3.4.21 &  3.5.1 & 3.5.2 & 3.6.1 & 3.6.4 & 3.6.5 & 4.1.1 & 4.2.1 & 4.6.1 & Avg  \\
\midrule
PROTSEED & -9.06&-9.79&-8.27&-9.02&-13.22&-9.77&-10.59&-6.70&-7.65&-10.32&-11.32&-13.50&-8.97&-5.74&-10.11&-9.63 \\
RFDiffusion+IF & -9.56&-6.81&-10.50&-10.00&-12.11&-12.35&-12.05&-9.66&-7.22&-13.16&-14.26&-11.10&-9.47&-5.92&-9.84&-10.51  \\
ESM2+EGNN &-11.30&-10.19&-10.85&-10.24&-13.65&-12.42&-12.80&-8.80&-9.61&-12.59&-15.68&-13.29&-10.36&-10.35&-14.61&-11.68 \\
\rowcolor{myblue}\model & -11.34&-13.10&-11.36&-12.66&-12.79&-15.25&-11.95&-11.15&-13.08&-12.24&-13.93&-14.17&-11.60&-9.34&-11.76&\textbf{-12.61} \\
\bottomrule
\end{tabular}
\vspace{-0.6em}
\caption{Substrate binding affinity~(kcal/mol, $\downarrow$) for the top-1 candidate. IF denotes the inverse folding model ProteinMPNN. Avg denotes average.}
\label{Tab: docking_top1}
\end{table*}

\begin{table*}[!t]
\tiny
\centering
\setlength{\tabcolsep}{0.6mm}
\begin{tabular}{lcccccccccccccccc}
\midrule
Enzyme Family & 1.1.1.239&1.1.1.270&1.1.1.201&1.1.1.184&1.1.1.25&1.1.1.62&1.1.1.197&1.1.1.35&1.1.1.248&1.1.1.271&1.1.1.372&1.1.1.357&1.1.1.27&1.1.1.286&1.1.1.47&1.1.1.3 \\
\rowcolor{myblue}
\model & 92.18&71.92&97.56&94.23&96.24&90.45&92.13&96.45&77.26&94.31&88.15&94.42&88.4&93.72&97.8&95.54 \\
\midrule
Enzyme Family & 1.1.1.56&1.1.1.105&1.1.1.77&1.1.1.34&1.1.1.399&1.1.1.42&1.1.1.49&1.1.1.44&1.1.1.40&1.11.1.24&1.11.1.12&1.11.1.15&1.11.1.9&1.11.1.11&1.11.1.5&1.11.1.6 \\
\rowcolor{myblue}
\model & 86.04&96.22&98.32&96.49&96.1&92.63&96.29&96.9&96.35&94.71&95.63&78.96&69.85&96.01&96.04&95.77\\
\midrule
Enzyme Family & 1.11.1.21&1.14.13.39&1.14.13.9&1.14.13.23&1.14.14.18&1.14.14.47&1.14.14.25&1.14.14.154&1.14.14.1&1.14.14.32&1.14.14.24&1.14.14.26&1.14.14.91&1.2.1.3&1.2.1.50&1.2.1.44  \\
\rowcolor{myblue}
\model & 93.84&93.16&90.26&95.01&90.42&96.8&96.37&94.63&88.46&96.79&89.97&95.54&80.31&93.12&40.98&73.57\\
\midrule
Enzyme Family & 1.2.1.12&1.2.1.59&1.2.1.13&1.2.1.11&1.2.1.10&1.2.1.5&1.2.1.47&1.2.1.36&1.2.1.9&1.2.1.26&1.2.1.88&1.2.1.84&1.2.1.89&1.2.1.90&2.1.1.309&2.1.1.37 \\
\rowcolor{myblue}
\model &97.54&84.78&88.9&96.02&89.11&98.09&95.01&95.72&97.27&97.94&92.61&29.82&90.8&95.95&67.58&58.33 \\
\midrule
Enzyme Family & 2.1.1.364&2.1.1.221&2.1.1.348&2.1.1.320&2.1.1.6&2.1.1.366&2.1.1.367&2.1.1.244&2.1.1.359&2.1.1.72&2.1.1.220&2.1.1.45&2.1.1.98&2.1.1.79&2.1.1.8&2.1.1.233 \\
\rowcolor{myblue}
\model &95.73&36.24&93.46&79.4&82.73&95.49&93.45&82.57&88.35&95.56&70.41&97.37&97.51&86.87&33.08&40.74\\
\midrule
Enzyme Family & 2.1.1.319&2.1.1.5&2.1.1.354&2.1.1.173&2.3.1.48&2.3.1.57&2.3.1.82&2.3.1.258&2.3.1.87&2.3.1.108&2.3.1.39&2.3.1.286&2.3.1.94&2.3.1.117&2.3.1.180&2.3.1.16 \\
\rowcolor{myblue}
\model &94.27&74.91&96.81&86.12&85.3&91.39&94.01&95.19&92.16&94.57&83.54&80.12&66.06&89.17&90.06&90.08\\
\midrule
Enzyme Family & 2.3.1.97&2.3.1.9&2.3.1.47&2.3.1.26&2.3.1.50&2.3.1.37&2.4.1.186&2.4.1.135&2.4.1.38&2.4.1.255&2.4.1.143&2.4.1.264&2.4.1.182&2.4.1.345&2.4.1.115&2.4.1.15 \\
\rowcolor{myblue}
\model &93.78&96.91&95.27&90.74&95.23&97.45&87.95&92.44&94.95&94.12&89.15&97.0&93.14&89.82&94.9&92.29\\
\midrule
Enzyme Family & 2.4.1.21&2.4.1.41&2.4.1.155&2.4.1.11&2.4.1.225&2.4.1.1&2.4.2.30&2.4.2.2&2.4.2.8&2.4.2.3&2.4.2.1&2.4.2.19&2.4.2.17&2.4.2.21&2.4.2.64&2.4.2.14 \\
\rowcolor{myblue}
\model &87.5&88.72&87.26&95.22&90.59&94.47&85.37&57.66&86.97&94.28&89.57&94.67&87.78&97.23&84.38&56.9 \\
\midrule
Enzyme Family & 2.4.2.36&2.5.1.18&2.5.1.31&2.5.1.68&2.5.1.55&2.5.1.16&2.5.1.1&2.5.1.15&2.5.1.79&2.5.1.87&2.5.1.29&2.5.1.59&2.5.1.126&2.5.1.47&2.5.1.46&2.5.1.54 \\
\rowcolor{myblue}
\model &95.83&90.4&93.98&90.52&96.3&96.63&94.58&87.21&84.97&48.7&93.43&96.64&93.75&97.17&95.03&92.67\\
\midrule
Enzyme Family & 2.5.1.6&2.5.1.7&2.5.1.19&2.5.1.26&2.6.1.21&2.6.1.9&2.6.1.16&2.6.1.99&2.6.1.51&2.6.1.63&2.6.1.3&2.6.1.1&2.6.1.57&2.6.1.7&2.6.1.81&2.6.1.73 \\
\rowcolor{myblue}
\model &85.44&95.85&97.85&94.61&98.39&95.83&92.46&33.65&87.92&97.82&92.37&91.69&63.4&97.38&92.31&94.62\\
\midrule
Enzyme Family & 2.6.1.83&2.6.1.62&2.6.1.85&2.6.1.82&2.7.1.107&2.7.1.154&2.7.1.26&2.7.1.71&2.7.1.21&2.7.1.24&2.7.1.113&2.7.1.153&2.7.1.23&2.7.1.35&2.7.1.33&2.7.1.68 \\
\rowcolor{myblue}
\model &97.21&84.11&65.48&91.0&82.46&80.82&34.34&90.39&86.82&94.29&88.44&85.81&84.82&95.64&90.21&92.01\\
\midrule
Enzyme Family & 2.7.1.20&2.7.1.178&2.7.1.19&2.7.1.159&2.7.1.82&2.7.1.73&2.7.1.40&2.7.1.1&2.7.1.47&2.7.1.158&2.7.1.5&2.7.10.2&2.7.10.1&2.7.11.1&2.7.11.21&2.7.11.25 \\
\rowcolor{myblue}
\model &95.98&84.15&89.49&37.84&82.38&38.89&95.48&89.56&96.63&94.46&96.0&88.84&85.6&87.22&93.53&88.0\\
\midrule
Enzyme Family & 2.7.11.22&2.7.11.30&2.7.11.24&2.7.11.26&2.7.11.2&2.7.11.16&2.7.4.6&2.7.4.3&2.7.4.8&2.7.4.14&2.7.4.9&2.7.4.25&2.7.4.10&2.7.4.22&2.7.4.24&2.7.4.1 \\
\rowcolor{myblue}
\model &89.9&87.49&82.13&95.35&87.32&93.13&96.78&89.4&92.22&93.66&90.56&91.12&76.71&92.77&94.7&72.34\\
\midrule
Enzyme Family & 2.7.7.7&2.7.7.6&2.7.7.65&2.7.7.60&2.7.7.86&2.7.7.31&2.7.7.12&2.7.7.52&2.7.7.27&2.7.7.23&2.7.7.9&2.7.7.48&3.1.1.4&3.1.1.3&3.1.1.78&3.1.1.2 \\
\rowcolor{myblue}
\model &89.35&70.58&59.94&95.05&63.21&95.83&92.19&31.92&93.16&31.28&88.78&97.82&94.83&88.53&54.0&89.44\\
\midrule
Enzyme Family & 3.1.1.23&3.1.1.72&3.1.1.47&3.1.1.34&3.1.1.8&3.1.1.7&3.1.1.5&3.1.3.32&3.1.3.46&3.1.3.3&3.1.3.104&3.1.3.16&3.1.3.108&3.1.3.7&3.1.3.11&3.1.3.36 \\
\rowcolor{myblue}
\model &81.44&98.51&93.26&96.08&94.52&96.32&26.47&50.77&92.35&94.74&42.83&82.73&70.04&78.52&88.95&90.43\\
\midrule
Enzyme Family & 3.1.3.10&3.1.3.76&3.1.3.1&3.1.4.52&3.1.4.17&3.1.4.55&3.1.4.35&3.1.4.53&3.2.2.6&3.2.2.9&3.2.2.16&3.2.2.8&3.2.2.3&3.2.2.1&3.2.2.10&3.4.19.12 \\
\rowcolor{myblue}
\model &88.66&92.35&97.53&82.9&88.1&38.32&96.68&92.31&77.62&85.88&33.32&95.85&94.01&66.52&57.02&65.74\\
\midrule
Enzyme Family & 3.4.19.5&3.4.19.14&3.4.19.13&3.4.21.53&3.4.21.92&3.4.21.107&3.4.21.22&3.5.1.98&3.5.1.26&3.5.1.97&3.5.1.124&3.5.1.9&3.5.1.44&3.5.1.1&3.5.1.2&3.5.1.25\\
\rowcolor{myblue}
\model &77.49&92.47&89.32&85.78&90.28&91.91&95.52&87.98&88.62&94.71&90.08&45.22&71.21&95.29&96.26&95.9\\
\midrule
Enzyme Family & 3.5.1.16&3.5.1.92&3.5.1.99&3.5.1.62&3.5.1.23&3.5.2.17&3.5.2.6&3.5.2.10&3.6.1.12&3.6.1.23&3.6.1.55&3.6.1.56&3.6.1.9&3.6.1.60&3.6.1.1&3.6.1.72 \\
\rowcolor{myblue}
\model &48.52&95.56&97.24&88.51&32.35&32.7&90.82&97.73&66.37&91.28&87.33&84.84&93.72&85.65&97.47&42.75\\
\midrule
Enzyme Family & 3.6.1.15&3.6.1.58&3.6.1.8&3.6.1.22&3.6.1.27&3.6.1.29&3.6.1.5&3.6.4.6&3.6.4.12&3.6.4.13&3.6.4.10&3.6.5.2&3.6.5.1&3.6.5.5&4.1.1.23&4.1.1.85\\
\rowcolor{myblue}
\model &77.06&95.34&61.92&78.6&73.23&57.51&78.07&77.19&69.1&79.55&79.61&84.59&86.29&69.46&84.22&95.33\\
\midrule
Enzyme Family & 4.1.1.17&4.1.1.37&4.1.1.39&4.1.1.28&4.1.1.12&4.1.1.49&4.1.1.25&4.1.1.32&4.2.1.10&4.2.1.19&4.2.1.1&4.2.1.24&4.2.1.22&4.2.1.8&4.2.1.2&4.6.1.1\\
\rowcolor{myblue}
\model &77.3&78.75&96.5&95.62&91.49&93.31&95.01&82.3&92.92&88.36&91.39&96.7&93.19&97.28&92.94&86.06\\
\midrule
Enzyme Family & 4.6.1.17&4.6.1.12&4.6.1.2&&&&&&&&&&&&&Avg\\
\rowcolor{myblue}
\model &67.84&95.33&87.6&&&&&&&&&&&&&87.45\\
\bottomrule
\end{tabular}
\vspace{-0.6em}
\caption{AlphaFold2 pLDDT~($\uparrow$) for the 323 testing fourth-level categories. Avg denotes average.}
\label{Tab: plddt_299families}
\end{table*}

\section{Additional Experimental Results}
\subsection{\model Performance on 323 Testing Fourth-Level Categories}
\label{appendix_299_family_results}
The ESP scores, substrate binding affinities and AlphaFold2 pLDDT results for the 323 testing fourth-level categories are respectively provided in Table~\ref{Tab: esp_299families},~\ref{Tab: docking_299families} and~\ref{Tab: plddt_299families}.


\subsection{Performance on Different Candidate Set}
\label{Appendix_different_candidate}
\subsubsection{Top-1 Candidate Results}
\label{Appendix: top1_results}
ESP scores of top-1 candidate and their corresponding substrate binding affinities as well as pLDDT are respectively reported in Table~\ref{Tab: ESP_top1}, ~\ref{Tab: docking_top1} and ~\ref{Tab: plddt_top1}. The tables show that our \model achieves the best average scores across different categories on all metrics. Notably, the ESP scores of different models consistently surpass those of the top-5 candidate. This observation is reasonable, as candidates with higher probabilities are more likely to achieve higher function scores.

\begin{table*}[!t]
\scriptsize
\centering
\setlength{\tabcolsep}{1.65mm}
\begin{tabular}{lcccccccccccccccc}
\midrule
Enzyme Family & 1.1.1 & 1.11.1 & 1.14.13 & 1.14.14 & 1.2.1 & 2.1.1 & 2.3.1 & 2.4.1 & 2.4.2 & 2.5.1 & 2.6.1 & 2.7.1 & 2.7.10 & 2.7.11 & 2.7.4 & Avg  \\
\midrule
PROTSEED &  86.28&82.47&73.36&86.48&91.38&64.90&91.83&74.33&89.81&74.43&72.72&79.61&78.34&70.53&92.71 & -- \\
RFDiffusion+IF & 83.88&97.98&92.53&84.18&88.08&84.90&76.39&98.00&71.68&76.93&76.79&97.88&77.24&95.80&81.25 &--   \\
ESM2+EGNN &96.35&97.19&88.50&92.12&96.64&77.03&95.73&94.09&94.69&96.40&96.44&93.70&94.43&92.31&97.50 &--\\
\rowcolor{myblue}
\model & 96.49&97.81&95.67&98.20&98.60&95.38&95.16&94.18&94.88&96.80&98.49&84.15&92.73&96.03&95.67 & \cellcolor{myblue}--  \\
\midrule
Enzyme Family & 2.7.7 & 3.1.1 & 3.1.3 & 3.1.4 & 3.2.2 & 3.4.19 & 3.4.21 & 3.5.1 & 3.5.2 & 3.6.1 & 3.6.4 & 3.6.5 & 4.1.1 & 4.2.1 & 4.6.1 & Avg  \\
\midrule
PROTSEED & 92.04&77.49&79.17&80.26&76.65&96.63&74.94&79.85&66.33&80.00&82.03&86.16&83.35&85.18&79.94&80.97 \\
RFDiffusion+IF & 87.40&85.01&70.33&73.15&96.12&94.79&77.33&87.29&54.46&83.95&81.33&69.30&82.97&82.47&78.36&82.93  \\
ESM2+EGNN &93.47&97.15&89.00&81.39&96.87&24.73&79.01&97.69&98.18&42.35&88.55&26.98&90.32&93.36&35.89&84.60 \\
\rowcolor{myblue}\model & 97.82&98.51&91.55&41.93&97.35&85.52&97.30&98.55&91.58&77.43&96.18&95.08&74.63&97.86&68.76&\textbf{91.34}\\
\bottomrule
\end{tabular}
\vspace{-0.6em}
\caption{AlphaFold2 pLDDT~($\uparrow$) for the top-1 candidate. IF denotes the inverse folding model ProteinMPNN. Avg denotes average.}
\label{Tab: plddt_top1}
\end{table*}

\subsubsection{Top-5 Candidate Results}
\label{Appendix: top5_results}

\begin{table*}[!t]
\scriptsize
\centering
\setlength{\tabcolsep}{1.8mm}
\begin{tabular}{lcccccccccccccccc}
\midrule
Enzyme Family & 1.1.1 & 1.11.1 & 1.14.13 & 1.14.14 & 1.2.1 & 2.1.1 & 2.3.1 & 2.4.1 & 2.4.2 & 2.5.1 & 2.6.1 & 2.7.1 & 2.7.10 & 2.7.11 & 2.7.4 & Avg  \\
\midrule
PROTSEED &  0.94&0.67&0.41&0.98&0.89&0.99&0.93&0.80&0.98&0.98&0.98&0.98&0.99&0.99&0.95&  -- \\
RFDiffusion+IF & 0.77&0.95&0.54&0.52&0.85&0.62&0.64&0.63&0.77&0.75&0.86&0.75&0.56&0.93&0.95 & -- \\
ESM2+EGNN &0.94&0.97&0.65&0.97&0.97&0.97&0.96&0.96&0.97&0.96&0.97&0.97&0.98&0.98&0.97 & -- \\
\rowcolor{myblue}
\model & 0.97&1.00&0.66&1.00&1.00&1.00&0.99&0.97&1.00&0.99&1.00&1.00&1.00&1.00&1.00 \\
\midrule
Enzyme Family & 2.7.7 & 3.1.1 & 3.1.3 & 3.1.4 & 3.2.2 & 3.4.19 & 3.4.21 &  3.5.1 & 3.5.2 & 3.6.1 & 3.6.4 & 3.6.5 & 4.1.1 & 4.2.1 & 4.6.1 & Avg  \\
\midrule
PROTSEED & 0.98&0.78&0.91&0.97&0.92&0.81&0.98&0.85&0.20&0.50&0.97&0.39&0.91&0.99&0.96&0.85 \\
RFDiffusion+IF & 0.88&0.57&0.89&0.95&0.70&0.95&0.59&0.76&0.58&0.85&0.83&0.84&0.91&0.95&0.81&0.77  \\
ESM2+EGNN &0.97&0.97&0.92&0.93&0.91&0.87&0.97&0.92&0.32&0.96&0.97&0.79&0.91&0.98&0.97&0.92\\
\rowcolor{myblue}
\model & 1.00&1.00&1.00&1.00&0.94&0.88&1.00&0.98&0.34&0.99&1.00&0.78&0.99&1.00&0.99&\textbf{0.95}\\
\bottomrule
\end{tabular}
\vspace{-0.6em}
\caption{Enzyme-substrate ESP score~($\uparrow$) for the top-5 candidate. IF denotes the inverse folding model ProteinMPNN. Note 0.6 or higher ESP score indicates a positive binding. Avg denotes average.}
\label{Tab: ESP_top5}
\end{table*}

ESP scores of top-5 candidate and their corresponding substrate binding affinities as well as pLDDT are respectively reported in Table~\ref{Tab: ESP_top5}, ~\ref{Tab: docking_top5} and ~\ref{Tab: plddt_top5}. The tables show that our \model achieves the best average scores across different categories on all metrics.

\begin{table*}[!t]
\scriptsize
\centering
\setlength{\tabcolsep}{1.60mm}
\begin{tabular}{lcccccccccccccccc}
\midrule
Enzyme Family & 1.1.1 & 1.11.1 & 1.14.13 & 1.14.14 & 1.2.1 & 2.1.1 & 2.3.1 & 2.4.1 & 2.4.2 & 2.5.1 & 2.6.1 & 2.7.1 & 2.7.10 & 2.7.11 & 2.7.4 & Avg  \\
\midrule
PROTSEED &  -5.99&-5.93&-5.55&-11.17&-6.64&-9.61&-11.83&-8.80&-11.84&-8.57&-6.01&-8.72&-11.30&-12.38&-12.59 &-- \\
RFDiffusion+IF & -7.03&-6.43&-6.70&-13.24&-8.42&-12.59&-11.36&-10.01&-12.74&-9.80&-7.99&-11.90&-12.30&-12.82&-12.90 & --   \\
ESM2+EGNN &-6.85&-8.46&-7.82&-14.01&-6.97&-12.04&-12.65&-11.80&-15.14&-9.34&-5.87&-10.01&-12.65&-12.20&-11.43 & -- \\
\rowcolor{myblue}\model & -14.11&-5.63&-6.71&-11.02&-10.05&-10.93&-12.66&-11.68&-13.94&-14.41&-9.14&-12.43&-12.70&-13.80&-12.47 \\
\midrule
Enzyme Family & 2.7.7 & 3.1.1 & 3.1.3 & 3.1.4 & 3.2.2 & 3.4.19 & 3.4.21 &  3.5.1 & 3.5.2 & 3.6.1 & 3.6.4 & 3.6.5 & 4.1.1 & 4.2.1 & 4.6.1 & Avg \\
\midrule
PROTSEED & -10.11&-7.00&-9.20&-9.68&-13.02&-10.12&-10.76&-5.53&-6.00&-10.57&-13.73&-11.76&-9.35&-6.74&-11.77&-9.41  \\
RFDiffusion+IF & -10.65&-9.15&-9.70&-10.11&-11.35&-13.10&-11.57&-6.88&-8.09&-11.07&-12.54&-11.45&-9.16&-6.64&-10.55&-10.27  \\
ESM2+EGNN &-10.23&-9.34&-9.77&-9.75&-13.48&-11.00&-11.27&-6.19&-12.08&-8.94&-13.61&-13.25&-9.69&-6.56&-10.84&-10.44 \\
\rowcolor{myblue}\model & -9.76&-10.73&-9.97&-12.42&-12.08&-11.58&-11.02&-9.51&-11.09&-11.58&-13.83&-12.90&-11.41&-6.52&-10.64&\textbf{-11.22} \\
\bottomrule
\end{tabular}
\vspace{-0.6em}
\caption{Substrate binding affinity~(kcal/mol, $\downarrow$) for the top-5 candidate. IF denotes the inverse folding model ProteinMPNN. Avg denotes average.}
\label{Tab: docking_top5}
\end{table*}

\begin{table*}[!t]
\scriptsize
\centering
\setlength{\tabcolsep}{1.65mm}
\begin{tabular}{lcccccccccccccccc}
\midrule
Enzyme Family & 1.1.1 & 1.11.1 & 1.14.13 & 1.14.14 & 1.2.1 & 2.1.1 & 2.3.1 & 2.4.1 & 2.4.2 & 2.5.1 & 2.6.1 & 2.7.1 & 2.7.10 & 2.7.11 & 2.7.4 & Avg  \\
\midrule
PROTSEED &  81.07&81.80&74.37&84.17&79.28&72.95&69.86&78.31&80.06&79.16&76.61&79.83&80.33&79.20&84.10 & -- \\
RFDiffusion+IF & 82.80&93.51&81.12&81.17&82.44&78.82&88.36&83.53&82.32&94.44&78.41&81.93&87.38&81.50&82.43 &--   \\
ESM2+EGNN &93.35&91.25&90.00&87.68&95.28&69.96&91.43&86.50&86.93&91.76&93.92&89.53&90.64&88.98&94.51&--\\
\rowcolor{myblue}
\model & 92.85&92.68&90.25&98.04&98.29&82.15&93.39&93.34&95.11&75.91&98.22&76.31&91.46&95.70&94.52 & \cellcolor{myblue}--  \\
\midrule
Enzyme Family & 2.7.7 & 3.1.1 & 3.1.3 & 3.1.4 & 3.2.2 & 3.4.19 & 3.4.21 &  3.5.1 & 3.5.2 & 3.6.1 & 3.6.4 & 3.6.5 & 4.1.1 & 4.2.1 & 4.6.1 & Avg  \\
\midrule
PROTSEED & 80.55&77.08&77.89&79.35&78.42&78.24&78.00&77.22&75.51&83.91&75.16&68.56&82.58&80.92&82.22&78.56 \\
RFDiffusion+IF & 81.95&81.99&76.11&81.04&92.59&89.93&78.36&87.08&79.25&87.46&60.63&88.13&75.89&77.25&80.03&82.60  \\
ESM2+EGNN &90.02&93.50&87.60&86.07&94.22&62.27&82.19&87.72&95.26&95.92&76.83&80.24&72.48&83.38&79.12&86.95 \\
\rowcolor{myblue}
\model & 96.80&98.04&89.22&81.27&96.82&71.04&96.64&98.12&80.57&80.78&95.23&80.05&87.88&93.38&90.87&\textbf{90.17}\\
\bottomrule
\end{tabular}
\vspace{-0.6em}
\caption{AlphaFold2 pLDDT~($\uparrow$) for the top-5 candidate. IF denotes the inverse folding model ProteinMPNN. Avg denotes average.}
\label{Tab: plddt_top5}
\end{table*}

\subsubsection{Top-10 Candidate Results}
\label{Appendix: top10_results}

\begin{table*}[!t]
\scriptsize
\centering
\setlength{\tabcolsep}{1.8mm}
\begin{tabular}{lcccccccccccccccc}
\midrule
Enzyme Family & 1.1.1 & 1.11.1 & 1.14.13 & 1.14.14 & 1.2.1 & 2.1.1 & 2.3.1 & 2.4.1 & 2.4.2 & 2.5.1 & 2.6.1 & 2.7.1 & 2.7.10 & 2.7.11 & 2.7.4 & Avg  \\
\midrule
PROTSEED &  0.90&0.67&0.38&0.98&0.78&0.99&0.82&0.55&0.98&0.96&0.96&0.98&0.99&0.99&0.80 &  -- \\
RFDiffusion+IF & 0.68&0.95&0.54&0.47&0.69&0.54&0.55&0.48&0.71&0.66&0.77&0.67&0.51&0.83&0.89& -- \\
ESM2+EGNN &0.89&0.96&0.59&0.94&0.93&0.96&0.91&0.88&0.95&0.95&0.94&0.95&0.96&0.97&0.95& -- \\
\rowcolor{myblue}
\model & 0.95&1.00&0.64&0.99&0.99&1.00&0.99&0.88&0.99&0.99&0.99&0.99&1.00&1.00&1.00 \\
\midrule
Enzyme Family & 2.7.7 & 3.1.1 & 3.1.3 & 3.1.4 & 3.2.2 & 3.4.19 & 3.4.21 &  3.5.1 & 3.5.2 & 3.6.1 & 3.6.4 & 3.6.5 & 4.1.1 & 4.2.1 & 4.6.1 & Avg  \\
\midrule
PROTSEED & 0.97&0.70&0.90&0.95&0.89&0.68&0.97&0.73&0.14&0.50&0.91&0.37&0.86&0.99&0.94&0.81 \\
RFDiffusion+IF & 0.78&0.49&0.85&0.89&0.55&0.92&0.47&0.67&0.53&0.73&0.66&0.76&0.84&0.95&0.60&0.69\\
ESM2+EGNN &0.95&0.95&0.90&0.90&0.84&0.62&0.94&0.86&0.23&0.93&0.95&0.71&0.89&0.97&0.93&0.88\\
\rowcolor{myblue}
\model & 0.99&0.99&0.99&1.00&0.91&0.69&1.00&0.94&0.27&0.98&1.00&0.74&0.99&1.00&0.99&\textbf{0.93}\\
\bottomrule
\end{tabular}
\vspace{-0.6em}
\caption{Enzyme-substrate ESP score~($\uparrow$) for the top-10 candidate. IF denotes the inverse folding model ProteinMPNN. Note 0.6 or higher ESP score indicates a positive binding. Avg denotes average.}
\label{Tab: ESP_10}
\end{table*}


\begin{table*}[!t]
\scriptsize
\centering
\setlength{\tabcolsep}{1.60mm}
\begin{tabular}{lcccccccccccccccc}
\midrule
Enzyme Family & 1.1.1 & 1.11.1 & 1.14.13 & 1.14.14 & 1.2.1 & 2.1.1 & 2.3.1 & 2.4.1 & 2.4.2 & 2.5.1 & 2.6.1 & 2.7.1 & 2.7.10 & 2.7.11 & 2.7.4 & Avg  \\
\midrule
PROTSEED &  -4.65&-4.93&-4.76&-11.09&-7.28&-9.46&-11.11&-9.21&-10.29&-7.88&-5.13&-8.38&-10.66&-11.37&-13.20 &-- \\
RFDiffusion+IF & -7.26&-5.43&-5.70&-11.15&-6.65&-11.45&-11.20&-9.33&-10.88&-8.84&-7.74&-10.02&-11.60&-11.66&-12.90 & --   \\
ESM2+EGNN &-5.38&-7.50&-6.75&-12.81&-8.16&-11.06&-10.54&-9.95&-13.16&-8.24&-5.53&-9.44&-11.88&-12.44&-10.84 & -- \\
\rowcolor{myblue}
\model & -10.49&-5.05&-5.82&-10.34&-8.18&-10.05&-10.38&-11.17&-11.91&-11.93&-8.84&-10.50&-11.81&-12.51&-11.50& \cellcolor{myblue}-- \\
\midrule
Enzyme Family & 2.7.7 & 3.1.1 & 3.1.3 & 3.1.4 & 3.2.2 & 3.4.19 & 3.4.21 &  3.5.1 & 3.5.2 & 3.6.1 & 3.6.4 & 3.6.5 & 4.1.1 & 4.2.1 & 4.6.1 & Avg  \\
\midrule
PROTSEED & -8.81&-7.01&-8.20&-8.71&-11.37&-9.50&-9.96&-5.21&-5.27&-13.57&-11.53&-10.37&-7.88&-5.24&-9.99&-8.60 \\
RFDiffusion+IF & -9.60&-7.51&-8.70&-8.82&-10.89&-11.41&-10.07&-6.43&-7.72&-10.07&-12.27&-10.87&-8.08&-6.14&-10.46&-9.36  \\
ESM2+EGNN &-8.83&-8.52&-8.92&-8.67&-9.73&-10.16&-10.60&-5.03&-8.67&-8.81&-10.89&-12.41&-8.53&-5.52&-9.45&-9.35 \\
\rowcolor{myblue}
\model & -9.17&-9.07&-9.62&-12.13&-11.26&-10.72&-10.69&-8.30&-8.66&-11.15&-13.10&-11.35&-11.29&-5.99&-10.64&\textbf{-10.12} \\
\bottomrule
\end{tabular}
\vspace{-0.6em}
\caption{Substrate binding affinity~(kcal/mol, $\downarrow$) for the top-10 candidate. IF denotes the inverse folding model ProteinMPNN. Avg denotes average.}
\label{Tab: docking_10}
\end{table*}

\begin{table*}[!t]
\scriptsize
\centering
\setlength{\tabcolsep}{1.65mm}
\begin{tabular}{lcccccccccccccccc}
\midrule
Enzyme Family & 1.1.1 & 1.11.1 & 1.14.13 & 1.14.14 & 1.2.1 & 2.1.1 & 2.3.1 & 2.4.1 & 2.4.2 & 2.5.1 & 2.6.1 & 2.7.1 & 2.7.10 & 2.7.11 & 2.7.4 & Avg  \\
\midrule
PROTSEED &  75.84&81.80&72.90&83.84&77.32&77.66&75.62&77.34&79.14&75.58&74.20&77.36&79.65&78.30&77.38 & -- \\
RFDiffusion+IF & 75.57&91.36&81.12&87.85&79.33&81.03&85.67&86.62&77.50&88.98&79.52&86.97&86.75&81.85&87.24 &--   \\
ESM2+EGNN &93.22&91.68&90.10&86.29&94.90&78.73&90.94&87.93&87.07&91.15&93.99&87.45&90.51&88.13&94.98 &--\\
\rowcolor{myblue}
\model & 90.05&91.30&92.56&97.56&98.08&89.81&88.91&93.67&91.56&86.30&97.75&73.66&90.53&94.81&96.02 & \cellcolor{myblue}--  \\
\midrule
Enzyme Family & 2.7.7 & 3.1.1 & 3.1.3 & 3.1.4 & 3.2.2 & 3.4.19 & 3.4.21 &  3.5.1 & 3.5.2 & 3.6.1 & 3.6.4 & 3.6.5 & 4.1.1 & 4.2.1 & 4.6.1 & Avg  \\
\midrule
PROTSEED & 76.90&76.29&77.89&79.41&77.11&72.06&79.91&73.83&79.13&75.16&68.74&83.46&76.33&78.37&85.71&77.47 \\
RFDiffusion+IF & 80.92&82.53&81.13&85.91&94.10&91.67&76.09&80.76&87.57&66.85&83.48&82.41&81.11&77.05&77.93&82.90\\
ESM2+EGNN &89.00&92.74&87.41&86.43&89.05&58.58&82.56&91.94&90.93&85.09&74.46&82.17&88.15&89.26&84.14&87.30\\
\rowcolor{myblue}
\model & 96.47&97.62&90.15&85.19&95.71&71.34&96.32&97.54&81.33&81.01&92.42&86.13&89.63&94.89&87.32&\textbf{90.52}\\
\bottomrule
\end{tabular}
\vspace{-0.6em}
\caption{AlphaFold2 pLDDT~($\uparrow$) for the top-10 candidate. IF denotes the inverse folding model ProteinMPNN. Avg denotes average.}
\label{Tab: plddt_10}
\end{table*}

\begin{table*}[ht]
\tiny
\begin{center}
\begin{tabular}{ll}
\midrule
PDB Source ID  & Sequence \\
\midrule
1KAG &  MLPPIFLVGPMGAGKTSVGRELARRLGLEFLDSDREIEERVGLTVAWIFEELGEANFRRREEEVLRELFSLEPPVLATGGGAVTNRKNREFLKRHGTVVYLEVPV \\
& EELLRRLRALPLLAKLEEKFRALFERLVALYRAAADIVVRNGLLVNLLVLLVR \\
5L2P & 
MPLPPHLEYIIQMMLEKGGKGFTTMSEVEEIRDSLLQSASNTEPVEVDIDKIKETFKTSYGVKARQYFPIDNKAYPVVLYLHGGSWVIGSKFTHDKVCRAITVSC \\
& NCKVISVDYRLAPEYKFPAAVYECYDATKWIYENAKKINIDITKIAIAGNSAGGNLAAAVSLLSKEKNIKLKYQILLYPAVSFDLDTKSYKEFADGHFLDTDMIKFI\\
&GNNYLFNSKEDKLNPYASPINFADLSNLPPALIVTAEYDPLRDVGEAYANKLLQAGVDVTSIRYKGLVHAFASVLDEIGDTINIMGKLLKEYFK\\
1L1E &MNLEERFFVLFLDPTQTYNCAYFERREMAEQEAQIAKIDAALGALGPEPGMTLLDIGCGWGATTRRLIEKYDVTVVGLDLAKEQANHVRQLLANSDNLRVRAA\\
&MLEKWDFFEEPVDRVVSIGAFEHFGYQDRAALFQKTKEHLKPDGGLLLHHIVVPKRREKQEQLLSPQHPLAKFKKLIKKEIFPGGNLPSYPQLMQAAEKAGWE\\
&VTREESLQLDGAYWWDWWATAAAAHADEAIAIQSEFAYEYQHEALAAAAGDFELGYWDIQIFVLQK \\
8DSO & HHHNSKSESLLLLKKKGLLGFGGKVFYGKWRGQYVVAVKKIRSQSSDESQFIEEVQVMQQLRHENLVQLYGVCTEGKDIYIITEYMAAGLLLYYVKKKKFQM\\
&QQILTICKAICSALEYLEEKTFLHHDLLKRNLLLNSEGVAKISDFGLAKFISSSSSTKEPFPPVVLEQSKFSSKSDIWAFGVLMWEEYSYGKMPFEKFTRQETLEHV\\
&AQGLRLYPPDIASEPVYKIMKECWHEKADERPSFSILLSNLEELQEELA \\
5TMP & RGKFIVIEGLDGAGKSTNIDVVVEQLQRDGRREEWVFTREPGGTPLAEKLRELLLTPDDEPDDVDMDTEMVLMFEAARSQLVETVIKPALARGAWVLCDRHDL\\
&STYAYQGGGRGLPVAKLQALESFAINGLRPDLTLYLDVPVEIGLRRAKQRGKLERFEQERFQQRVAAVYAGRLERAQQDDSWQTIDATQPLSTVSDAIRTHLQRL\\
&QSEL\\
6W7O & HHLKLTFLFKKGELGGFFGGVFYGKWRGQYEVAVKKIRKQSSSESQFIDEVQVMQQLSHENLVQLYGCCTEGDDIYIITEYMANGSLLQYYLKKQHQFQQSQL\\
&LQIAKDIVSALEYLEELTLHHPGLSKRNLLLNSEGVAKITDFGLSRRVLDSDISIISSSFSLLPPIPEEELQQSKFTSKSDIWSFGVLMWEEYSYGKMPWQRFSQQE\\
&MEEHIRQQLRLYPPDLASPPVYTIMKECWHEKADERPSFSILLSNLVLLLE\\
3UYO & GSTKSTSEESHSWYHGPIPREDAERLLVSGIHGSFLVRESESTAGDHSLSLFYEGVKYHYRINTALDGKLYVSEGIKFDTLAELVQHYKTHADGLCYVLSEPCPPV
\\&GEEEEEEE\\
4NI2 & QVIAQKHDNVSILFSDIVGFTSLASQCSAQDLVMTLNELFSRFDKLCAENDVYKIETIGDAYCCVSGLPRKEPDHSQQICEMALDMMEVSSQVKTPHGKPINMRI\\
&GIHSGRVVAGVVGLMRRYYCLFGNDVNLANHMESGGVPGKINVPKETYELLKNDFSIEPRRGGSEECPENFPKKIVGYCLFVRRAAA\\
\bottomrule
\end{tabular}
\end{center}
\caption{Enzyme sequences for all analyzed cases.}
\label{Tab: appendix_case_study_seq}
\end{table*}

ESP scores of top-10 candidate and their corresponding substrate binding affinities as well as pLDDT are respectively reported in Table~\ref{Tab: ESP_10}, ~\ref{Tab: docking_10} and ~\ref{Tab: plddt_10}. Again, our model achieves the best average performance in all cases. It is worth noting that our \model achieves an average enzyme-substrate ESP score of 0.93, an average substrate binding affinity of -10.12 and an average pLDDT of 90.52, demonstrating \model's ability to design both well-folded and functional enzymes. 

\subsection{More Novel Cases}
\label{Appendix: case_study}
\begin{figure*}
	\centering
	\subfigure[Complex of designed 1L1E and S-adenosyl-L-methionine zwitterion. pLDDT=86.86, binding affinity=-11.02, Uniprot blastp recovery rate=56.5\%]{
		\includegraphics[width=4.6cm]{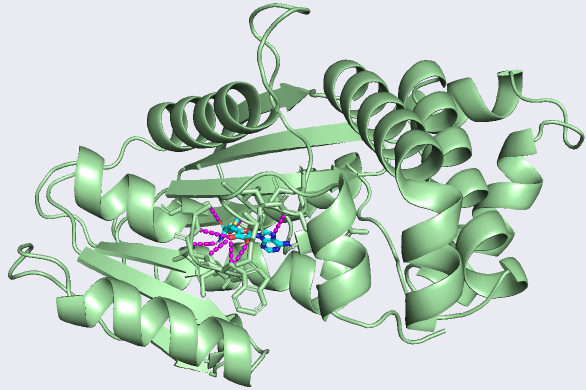}
	}
	\quad
 \subfigure[Complex of designed 8DSO and ATP. pLDDT=80.95, binding affinity=-11.19, Uniprot blastp recovery rate=58.1\%]{
		\includegraphics[width=4.2cm]{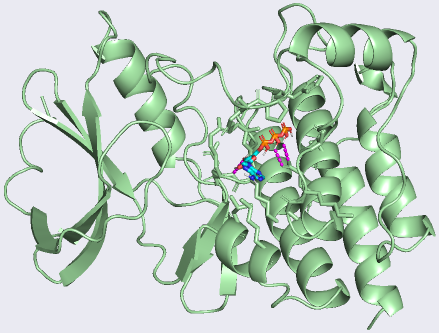}
	}
     \quad
     \subfigure[Complex of designed 5TMP and ATP(4-). pLDDT=85.34, binding affinity=-11.76, Uniprot blastp recovery rate=59.8\%]{
		\includegraphics[width=4.2cm]{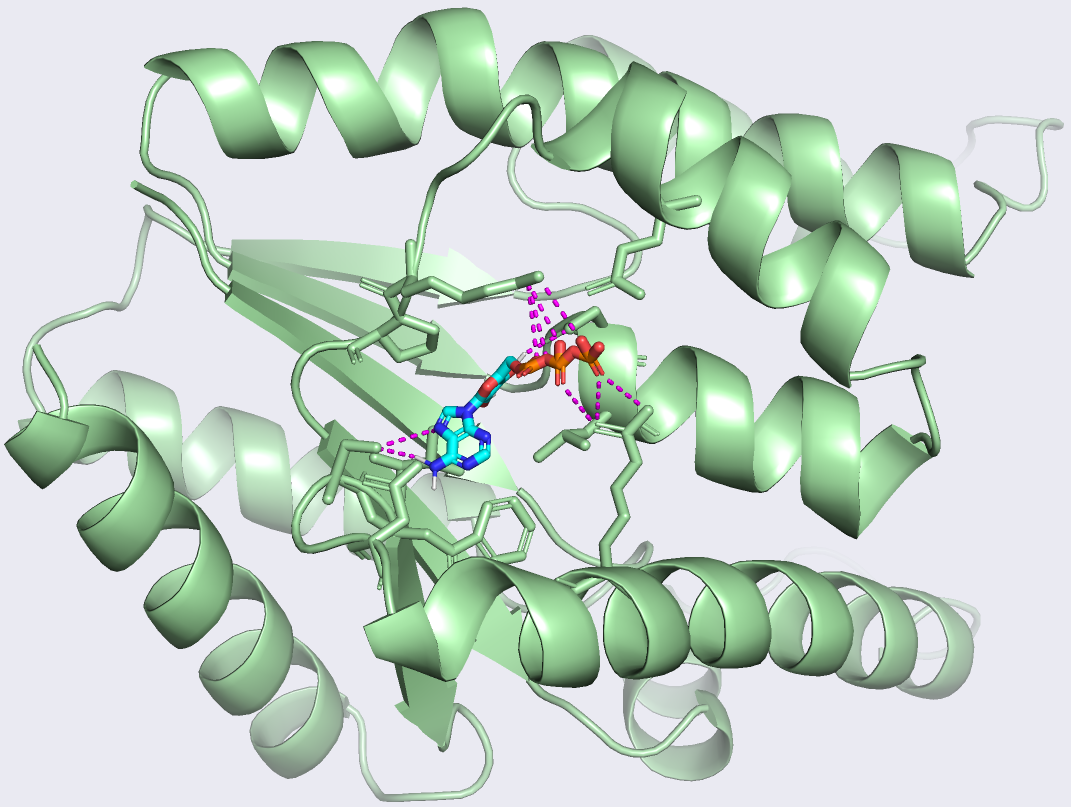}
	}
\quad
     \subfigure[Complex of designed 6W7O and ATP. pLDDT=80.84, binding affinity=-10.52, 
   Uniprot blastp recovery rate=58.0\%]{
		\includegraphics[width=5.3cm]{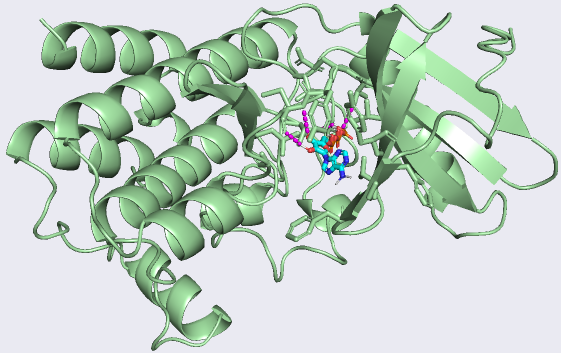}
	}
\quad
  \subfigure[Complex of designed 3UYO and ATP. pLDDT=81.69, binding affinity=-11.6, Uniprot blastp recovery rate=64.7\%]{
		\includegraphics[width=3.8cm]{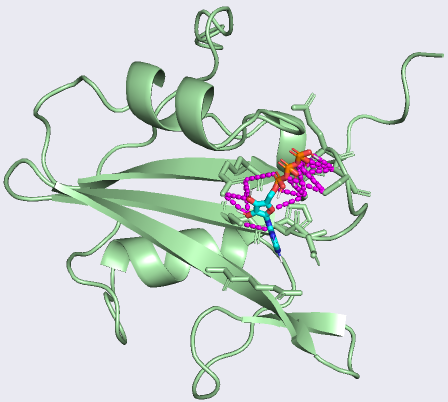}
	}
 \quad
 \subfigure[Complex of designed 4NI2 and ATP. pLDDT=81.32, binding affinity=-11.92, Uniprot blastp recovery rate=55\%]{
		\includegraphics[width=4.1cm]{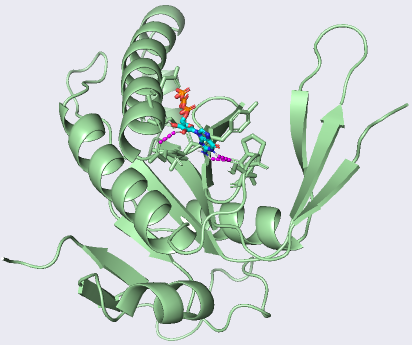}
	}
	\caption{Complexes of the \model designed enzymes and their corresponding substrates.}
	\label{Appendix_fig: case_study}
\end{figure*}

We provide more designed enzymes in Figure~\ref{Appendix_fig: case_study}. All these cases achieve a pLDDT higher than 80, a blastp amino acid identity rate lower than 65\% in Uniprot, and a substrate binding affinity lower than -10. After Gnina docking, the corresponding substrates demonstrate polar contacts~(hydrogen bonds) with the enzymes~(shown in purple), affirming the enzyme-substrate interaction functions. We also provide the enzyme sequences of all the analyzed cases in Table~\ref{Tab: appendix_case_study_seq}.

\end{document}